\documentclass[letterpaper,journal]{IEEEtran}

\usepackage{amsmath,amsfonts}
\usepackage{algorithm}
\usepackage{algpseudocode}
\usepackage{array}
\usepackage{tabularx}
\usepackage[caption=false,font=normalsize,labelfont=sf,textfont=sf]{subfig}
\usepackage{textcomp}
\usepackage{stfloats}
\usepackage{url}
\usepackage{verbatim}
\usepackage{graphicx}
\usepackage[nocompress]{cite}
\usepackage{booktabs}
\usepackage{multirow}
\usepackage{makecell}
\usepackage{amssymb}
\usepackage[table]{xcolor}
\usepackage[colorlinks,linkcolor=blue,anchorcolor=blue,citecolor=blue]{hyperref}
\usepackage{orcidlink}

\newcommand{\pubyear}[1]{{\textsubscript{#1}}}

\begin{document}
\begin{sloppypar}

\title{GeoDINO: A Unified Geometry-Aware Framework for Multimodal Remote Sensing Semantic Segmentation Across Domains}

\author{Shufan~Cheng$^{\orcidlink{0009-0007-1618-0770}}$, Huatao~Shang$^{\orcidlink{0000-0002-9969-4510}}$, Liang~Yin$^{\orcidlink{0000-0002-8344-5608}}$,~\IEEEmembership{Member,~IEEE,} Jiaxuan~Fang$^{\orcidlink{0009-0001-8184-7849}}$, Zequan~Wang$^{\orcidlink{0009-0008-4993-4804}}$,~\IEEEmembership{Graduate~Student~Member,~IEEE,} and Yitong~Liu$^{\orcidlink{0000-0002-1202-6174}}$,~\IEEEmembership{Member,~IEEE}

\thanks{Received 27 October 2026; accepted 12 November 2026. This work was supported in part by the National Key Laboratory of Space Target Awareness Fund and in part by the Beijing University of Posts and Telecommunications Excellent Ph.D. Students Foundation under Grant CX20260023. \textit{(Corresponding author: Liang~Yin.)}}
\thanks{Shufan~Cheng, Liang~Yin, Jiaxuan~Fang, Zequan~Wang, and Yitong~Liu are with the School of Information and Communication Engineering, Beijing University of Posts and Telecommunications, Beijing 100876, China (e-mail: csf@bupt.edu.cn; YinL@bupt.edu.cn; fangjiaxuan@bupt.edu.cn; zequanwang@bupt.edu.cn; liuyitong@bupt.edu.cn).}
\thanks{Huatao~Shang is with Beijing Institute of Radio Measurement, Beijing 100854, China (e-mail: shanghuatao\_bit@163.com).}
\thanks{Digital Object Identifier 10.1109/TGRS.2026.3755058}}

\markboth{IEEE TRANSACTIONS ON GEOSCIENCE AND REMOTE SENSING}{CHENG \MakeLowercase{\textit{et al.}}: G\MakeLowercase{eo}DINO: A UNIFIED GEOMETRY-AWARE FRAMEWORK}

\makeatletter\def\@IEEEpubidpullup{3.3\baselineskip}\makeatother
\IEEEpubid{\begin{minipage}{\textwidth}\centering\footnotesize 1558-0644 \textcopyright{} 2026 IEEE. All rights reserved, including rights for text and data mining, and training of artificial intelligence and similar technologies. Personal use is permitted, but republication/redistribution requires IEEE permission. See https://www.ieee.org/publications/rights/index.html for more information.\end{minipage}}

\maketitle

\begin{abstract}
  Optical  imagery and digital surface models (DSMs) provide complementary
appearance and geometric   cues for remote sensing  semantic segmentation.
In practice, pixel-wise labels are costly and cover only a few regions, so
segmentation trained on labeled regions has to be accurate both on these regions and on new
regions without labels.
However, existing   multimodal segmentation methods, including recent
foundation-model-based ones, focus mainly on in-domain accuracy and pay
less attention to generalization across domains, whereas  domain generalization (DG)   methods
target cross-domain generalization but are designed mainly for  single-modal
  optical imagery. To this end, we propose GeoDINO, a unified geometry-aware
and parameter-efficient framework   that learns from a single labeled source
domain and remains accurate  both in-domain and   on unseen target domains.
GeoDINO uses DSM geometry to guide a frozen  DINOv3 encoder: GeoPrior derives
pairwise structural relations from the DSM, and geometry-aware low-rank adaptation (GeoLoRA) injects them as an
attention bias  into the optical branch during parameter-efficient
fine-tuning (PEFT).   A  deformable sparse cross-attention (DeSCA) module   then
selects which cross-modal interactions are fused, and a  multimodal masked
image modeling (mmMIM) objective with random single-modality masking   trains
the model under incomplete observations. On five in-domain tasks
on ISPRS Vaihingen, ISPRS Potsdam, and MMHunan, GeoDINO
improves mean intersection over union (mIoU) by 1.20\% on average  over existing single-modal and multimodal
segmentation methods. On eight cross-domain tasks covering region shift,
spectral shift, and their combination,  it improves mIoU by 1.23\% on average
over   source-only domain-adaptation architectures, single-modal remote
sensing   DG  models, and same-framework PEFT variants.
\end{abstract}

\begin{IEEEkeywords}
Domain generalization (DG), multimodal fine-tuning, remote sensing, semantic segmentation, vision foundation model (VFM)
\end{IEEEkeywords}

\section{Introduction}

\IEEEPARstart{S}{emantic} segmentation is a core task in remote sensing image
understanding and supports a wide range of applications, including land-cover
mapping, urban planning, infrastructure monitoring, and disaster
response~\cite{HuangSurvey}. With the growing availability of heterogeneous
Earth observation data, multimodal remote sensing has become increasingly
important for scene understanding~\cite{DLMMRSReview}. For high-resolution urban scenes, optical
imagery and digital surface models (DSMs) form one of the most widely used and
effective multimodal combinations~\cite{Jin2022Fusion,vFuseNet}: optical data
provide rich appearance cues, while DSM supplies geometric structure for
disambiguating visually similar categories and sharpening object boundaries.
In practice, pixel-wise labels are costly to produce\mbox{\cite{DAFormer}} and cover only a few
regions, so segmentation trained on labeled regions has to serve both these regions and new
regions without labels.
However, most existing methods are still developed under the same-distribution
assumption, and their performance often degrades markedly when deployed across
regions, seasons, sensors, spectral settings, spatial resolutions, and
annotation protocols~\cite{LoveDA,CASID}. Achieving both strong in-domain
accuracy and robust cross-domain generalization for multimodal remote sensing
segmentation therefore remains an open challenge.

\IEEEpubidadjcol
Existing studies can be organized into two largely separate research lines.
The first line focuses on multimodal fusion for in-domain segmentation.
Early single-modal methods relied on CNN encoder--decoder frameworks such as
PSPNet~\cite{PSPNet} for multiscale context modeling. The introduction of vision transformers~\cite{ViT} and hierarchical variants such as the Swin Transformer~\cite{SwinTransformer} brought long-range dependency modeling to
dense prediction, leading to hybrid architectures like
TransUNet~\cite{TransUNet} and UNetFormer~\cite{UNetFormer} that combine
convolutional local detail with transformer global context. However,
single-modal optical features alone struggle with appearance ambiguity and
shadow occlusion in complex urban scenes, motivating multimodal optical--DSM
fusion. Driven by deep learning, multimodal segmentation has progressed from
hand-crafted fusion pipelines~\cite{Zhang2015DMTHP,Lefevre2019SegFusion} to
end-to-end dual-stream CNN architectures~\cite{vFuseNet,ESANet}, stronger
intermediate cross-modal interaction modules~\cite{CMGFNet,CMFNet}, and
transformer-based multimodal designs~\cite{FTransUNet,FTransDeepLab}. More
recently, multimodal fine-tuning of vision foundation models (VFMs) has further
advanced in-domain performance~\cite{MFNet}. Despite these advances, most
multimodal methods are still developed under a closed-world assumption in
which training and test data follow similar distributions, leaving
cross-domain robustness largely unaddressed.

\begin{figure}[!t]
\centering
  \includegraphics[width=\columnwidth]{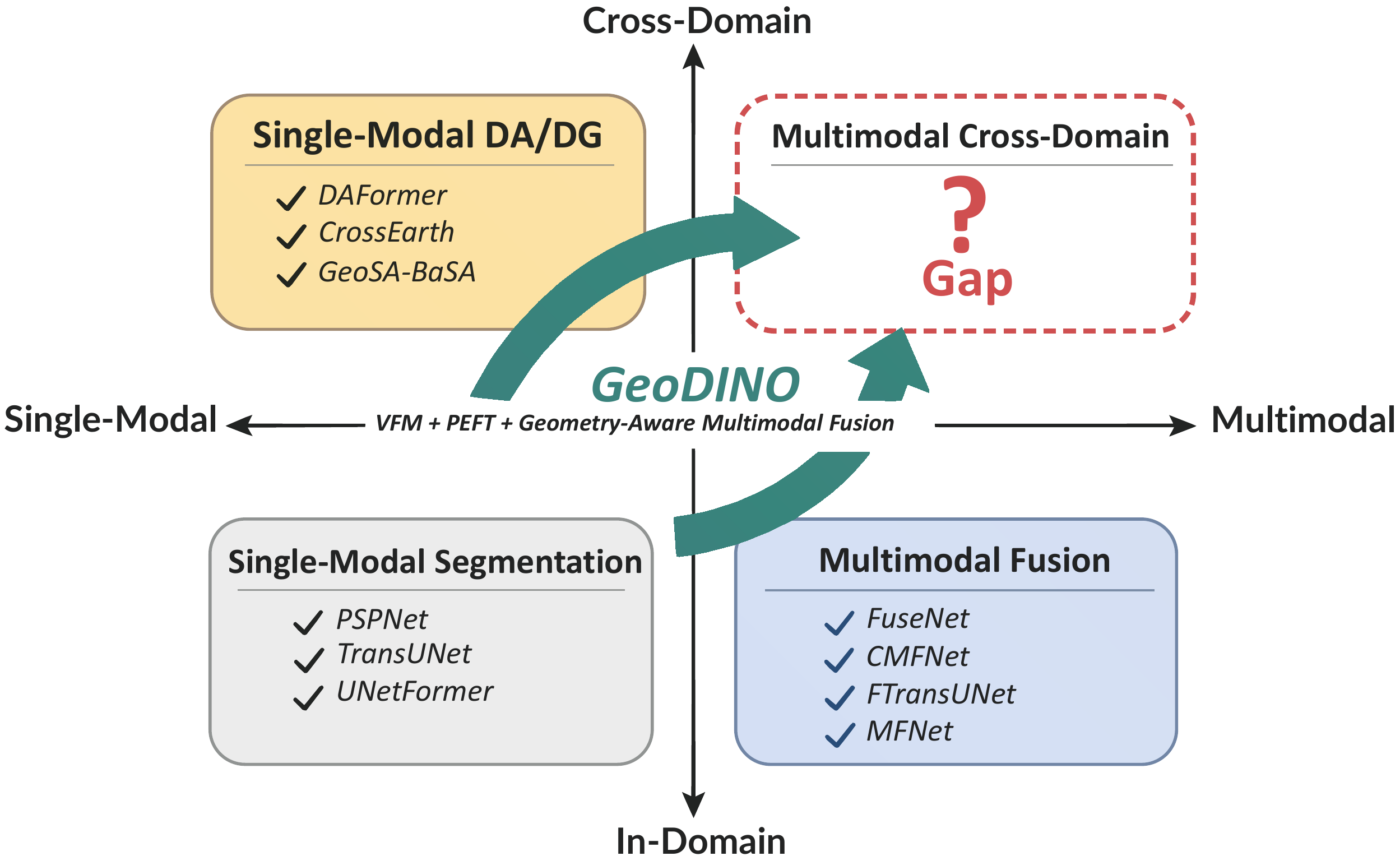}
 \caption{Two disconnected research lines in multimodal cross-domain remote
sensing segmentation. Multimodal fusion methods improve in-domain accuracy
but assume matched source--target distributions, while DA/DG
 methods address domain shift but operate on single-modal
optical imagery. GeoDINO bridges these two lines by combining multimodal
geometric cues with parameter-efficient   VFM  adaptation.}
\label{fig:intro_overview}
\end{figure}

The second line focuses on cross-domain remote sensing
segmentation. Domain adaptation (DA) methods~\cite{DAFormer,HRDA} improve transfer
performance when target-domain data are available during training, whereas
domain generalization (DG) methods, particularly single-modal remote sensing DG (RSDG) methods~\cite{FOSMix,CrossEarth,GeoSABaSA}, aim to
generalize without any target-domain samples. However, these RSDG methods are
designed predominantly for single-modal optical imagery and do not exploit
auxiliary modalities such as DSM, whose geometric cues could provide more
stable cross-domain priors. As illustrated in Fig.~\ref{fig:intro_overview},
these two lines remain disconnected: multimodal methods prioritize
cross-modal fusion but ignore domain shift, whereas DG methods address domain
shift but ignore multimodal information. For multimodal remote sensing, the
problem is further complicated by a two-axis shift: geometric distributions
vary with terrain, acquisition geometry, and DSM reconstruction quality,
while cross-modal correspondence degrades under misregistration, missing
measurements, and modality-specific noise. Conventional fusion mechanisms
therefore easily overfit source-domain cross-modal correlations and become
brittle under realistic distribution shifts.

Recently, VFMs and parameter-efficient fine-tuning
(PEFT) have been increasingly adopted to improve transferability in
remote sensing~\cite{WanSAMSurvey}. Pretrained on large-scale   natural  and
remote sensing   images, VFMs such as DINOv2~\cite{DINOv2}, SatMAE~\cite{SatMAE}, and RemoteCLIP~\cite{RemoteCLIP} have
shown stronger transferability than task-specific backbones in many
downstream settings, while PEFT methods such
as Adapter~\cite{Adapter}, low-rank adaptation (LoRA)~\cite{LoRA}, VPT~\cite{VPT}, and Rein~\cite{Rein} make VFM adaptation practical under limited
training budgets. These techniques have already
shown value for multimodal in-domain segmentation~\cite{MFNet} and for
single-modal DG~\cite{CrossEarth,GeoSABaSA}, yet a unified
framework that   uses  multimodal geometric cues   to guide VFM adaptation for
 both in-domain   accuracy and cross-domain generalization is still
lacking. To address this gap, we propose GeoDINO, a geometry-aware and
parameter-efficient framework for multimodal remote sensing semantic
segmentation   that learns from a single labeled source domain and remains
accurate  both in-domain and   on unseen target domains. Built on a shared DINOv3
encoder~\cite{DINOv3}, GeoDINO   uses DSM geometry to guide the adaptation of
the frozen encoder and fuses the two modalities through  selective
cross-modal interaction. Finally, we incorporate a multimodal masked image
modeling  (mmMIM)  objective with random single-modality masking to   further
improve cross-domain performance.

The main contributions of this work are summarized as follows.

\begin{enumerate}
    \item A unified geometry-aware framework, GeoDINO, is proposed to bridge
    in-domain multimodal semantic segmentation and cross-domain generalization in remote sensing.
    Learning from a single labeled source domain while remaining accurate
    under both matched and shifted test distributions is a requirement
    that multimodal fusion methods and single-modal DG methods each address
    only in part.

    \item A geometry-aware multimodal adaptation strategy is introduced that
    uses the strong pretrained features of a frozen DINOv3 encoder to maintain
    in-domain accuracy and limits the fitting of source-specific cues during
    single-source training. It acts at the three points where such fitting
    can occur: GeoPrior and geometry-aware LoRA (GeoLoRA) inject DSM-derived structural priors into
    PEFT as an attention bias that constrains how the encoder is adapted, a
    deformable sparse cross-attention (DeSCA) module selects which
    cross-modal interactions are fused, and mmMIM with random single-modality
    masking trains the model under incomplete observations to improve
    generalization to unseen domains.

    \item Extensive evaluations on Vaihingen, Potsdam, and MMHunan validate the
    effectiveness of GeoDINO. On five in-domain tasks, GeoDINO improves mean intersection over union (mIoU) by
    1.20\% on average over existing single-modal and multimodal segmentation
    methods; on eight cross-domain tasks covering region shift, spectral shift,
    and their combination, it improves mIoU by 1.23\% on average over
    source-only DA architectures, single-modal RSDG models, and same-framework
    PEFT variants.
\end{enumerate}

\section{Related Work}

\subsection{Remote Sensing Semantic Segmentation}
Semantic segmentation is a fundamental task in remote sensing image understanding. Early progress was
mainly driven by single-modality optical segmentation models based on encoder-decoder frameworks and
multiscale context modeling, such as PSPNet~\cite{PSPNet}. With the adoption of vision transformers~\cite{ViT,SwinTransformer},
architectures such as TransUNet~\cite{TransUNet} and UNetFormer~\cite{UNetFormer} further improved
dense prediction by modeling long-range spatial dependencies.  Frequency-domain
modeling has recently become another active line of remote sensing
segmentation. SFFNet\mbox{\cite{SFFNet}} fuses wavelet-domain and spatial features,
FDNet\mbox{\cite{FDNet}} decouples high- and low-frequency components and refines
them separately, and AFENet\mbox{\cite{AFENet}} adaptively separates and enhances
the two frequency bands according to image content. For general dense
prediction, FreqFusion\mbox{\cite{FreqFusion}} fuses multilevel features with
adaptive low- and high-pass filters.  However, optical imagery alone often
struggles in complex urban scenes under appearance ambiguity, shadow occlusion, and structural
similarity across categories. These limitations have driven increasing interest in multimodal remote
sensing segmentation with auxiliary modalities such as DSM.

Existing multimodal methods can be organized by the stage and strength of
cross-modal interaction. Early dual-stream models such as
FuseNet~\cite{FuseNet}, V-FuseNet~\cite{vFuseNet},
ESANet~\cite{ESANet}, and SA-GATE~\cite{SAGATE} mainly perform progressive,
late, or gated fusion on separately encoded optical and depth features. Later
remote-sensing-specific methods move toward stronger intermediate fusion:
CMGFNet~\cite{CMGFNet} adopts dual encoders with gated multilevel
aggregation, CMFNet~\cite{CMFNet} introduces multiscale cross-modal
attention, and MFTransNet~\cite{MFTransNet} combines convolutional feature
extraction, multihead fusion, and multiscale decoding in a lightweight
CNN--Transformer hybrid framework. FTransUNet~\cite{FTransUNet} and
FTransDeepLab~\cite{FTransDeepLab} further enhance multimodal interaction
through hierarchical CNN--Transformer fusion and multiscale context
modeling. More recent methods shift toward more unified multimodal
frameworks: MultiSenseSeg~\cite{MultiSenseSeg} supports flexible modality
combinations in a cost-effective manner, while MFNet~\cite{MFNet} couples
  multimodal fine-tuning of the Segment Anything Model (SAM)\mbox{\cite{SAM}}  with pyramid deep fusion, indicating a
growing trend toward   VFM-based  multimodal segmentation.
 Following this trend, MM-DINO\mbox{\cite{MMDINO}} keeps a DINOv3 backbone frozen
and performs early soft cross-modal interaction in a modality-adaptive
adapter, and MoBaNet\mbox{\cite{MoBaNet}} alleviates modality imbalance in the parameter-efficient multimodal
adaptation of a frozen VFM. Other recent studies refine the
cross-modal interaction itself. CMX\mbox{\cite{CMX}} rectifies the features of each
modality with those of the other before fusion, EG-CMFNet\mbox{\cite{EGCMFNet}}
combines spatial--channel feature interaction with an edge-guided auxiliary
branch for boundary-aware segmentation of optical and synthetic aperture radar
(SAR) images,
SiMultiF\mbox{\cite{SiMultiF}} adaptively allocates modality weights across scenes
on top of a Siamese encoder, LMFNet\mbox{\cite{LMFNet}} adopts a lightweight
weight-sharing multimodal transformer, and HDSMNet\mbox{\cite{HDSMNet}} uses height
cues to guide sparse fusion of optical imagery and normalized DSM.

Although these methods have substantially improved in-domain segmentation
accuracy, most of them are developed under the assumption that training and
test data share similar geographic and sensing distributions.
Consequently, the robustness of multimodal remote sensing segmentation in
cross-domain settings remains insufficiently explored.

\subsection{Domain Adaptation and Generalization in Remote Sensing}
Domain shift is a fundamental challenge in remote sensing because data
distributions vary across geographic regions, seasons, atmospheric
conditions, sensor types, spectral configurations, and spatial
resolutions. Existing cross-domain studies mainly follow two settings   that
differ in the data available during training. DA assumes
access to target-domain data, usually unlabeled, during training, whereas
DG learns from labeled source domains only and is
evaluated on target domains unseen during
training\mbox{\cite{ZhouDGSurvey,WangDGSurvey}}. In remote
sensing segmentation, DA has been supported by benchmarks such as
LoveDA~\cite{LoveDA}, WHU Building~\cite{WHUBuilding}, and
CASID~\cite{CASID}, which cover urban to rural, cross-source, and climate
shifts, as well as strong baselines such as
DAFormer~\cite{DAFormer} and HRDA~\cite{HRDA}. These studies show that
pseudo-label refinement and resolution-aware training can substantially
improve cross-domain performance when target-domain data are available,
but the DA paradigm fundamentally relies on such access.

Under the DG setting, existing methods mainly improve generalization by suppressing
domain-specific appearance variation or learning more robust structural representations.
IBN-Net~\cite{IBNNet} combines instance normalization and batch normalization to improve robustness,
PASTA~\cite{PASTA} perturbs the Fourier amplitude spectrum to diversify source styles, and
HGFormer~\cite{HGFormer} strengthens domain-generalized segmentation through hierarchical
part-to-whole grouping. In remote sensing, this line of research is still relatively limited.
FOSMix~\cite{FOSMix} extends frequency-based augmentation to remote sensing by combining style
mixing in the frequency domain with consistency regularization under regional and resolution shifts.
Nevertheless, these methods still rely primarily on source-domain enhancement through appearance
perturbation or auxiliary-domain generation.

Current cross-domain remote sensing studies focus mainly on single-modal
  optical imagery, while DG for multimodal semantic segmentation, in which a
model trained on a labeled source domain with optical and DSM inputs is
applied to unseen target domains,  remains
largely unexplored. Recent studies such as
CrossEarth~\cite{CrossEarth}, GeoSA-BaSA~\cite{GeoSABaSA}, and
Earth-Adapter~\cite{EarthAdapter} have started to investigate   VFMs  and efficient adaptation strategies for cross-domain
remote sensing.  D$^2$FT-Net\mbox{\cite{D2FTNet}} further fine-tunes VFMs jointly
in the frequency and spatial domains for
domain-generalized remote sensing segmentation, and FADA\mbox{\cite{FADA}} adapts
the low- and high-frequency components of frozen VFM features
separately for domain-generalized segmentation of natural images.  These works suggest that large
pretrained priors and lightweight adaptation may provide a promising direction for improving
generalization under domain shift.

\subsection{Vision Foundation Models and Efficient Fine-Tuning}
  VFMs  are increasingly used for remote sensing
representation learning and can be grouped into several families. In the
vision--language direction, CLIP-style models~\cite{CLIP} align images with text, and
RemoteCLIP~\cite{RemoteCLIP} extends this paradigm to remote sensing by
constructing large-scale image--text pairs from heterogeneous annotations.
In the generative self-supervised direction, masked image modeling (MIM) is
a major route for remote sensing pretraining. SatMAE~\cite{SatMAE} adapts
masked autoencoders (MAEs)~\cite{MAE} to temporal and multispectral satellite imagery,
SatMAE++~\cite{SatMAEPP} further introduces multiscale pretraining for
multispectral data, and RingMo~\cite{RingMo} develops a remote sensing
foundation model with large-scale pretraining data and a masking strategy
tailored to dense scenes with abundant small objects. For specific modalities
and settings, HyperSIGMA~\cite{HyperSIGMA} targets hyperspectral image
understanding, while SkySense~\cite{SkySense} builds a large-scale multimodal
spatiotemporal remote sensing foundation model with multigranularity
contrastive learning and geo-context prototype learning, and
SkySenseV2~\cite{SkySenseV2} further moves toward a unified multimodal
transformer with adaptive patch merging, modality prompts, and
mixture-of-experts design.

The DINO family forms another important line of VFMs. From DINO~\cite{DINO}
to DINOv2~\cite{DINOv2}, DINO with Registers~\cite{DINORegister}, and
DINOv3~\cite{DINOv3}, these models show strong transferability for dense
prediction and segmentation. CrossEarth~\cite{CrossEarth} and
CrossEarth-SAR~\cite{CrossEarthSAR} demonstrate the use of DINO-style
pretrained representations in cross-domain geospatial segmentation, while
ChangeDINO~\cite{ChangeDINO} and Dino U-Net~\cite{DinoUNet} show that
DINOv3 features are also effective for change detection and medical image
segmentation. The SAM~\cite{SAM} family provides another route
for   VFM  transfer. RingMo-SAM~\cite{RingMoSAM},
RemoteSAM~\cite{RemoteSAM}, and
ClassWise-SAM-Adapter~\cite{ClasswiseSAMAdapter} adapt promptable
segmentation to multimodal remote sensing, Earth observation imagery, and
SAR semantic segmentation, respectively.

For downstream adaptation, existing strategies mainly include freezing the
pretrained backbone, full fine-tuning, and   PEFT. Typical PEFT methods include Adapter~\cite{Adapter},
LoRA~\cite{LoRA}, VPT~\cite{VPT}, and AdaptFormer~\cite{AdaptFormer}.
Rein~\cite{Rein} further shows that lightweight feature adaptation can
unlock the generalization potential of VFMs in dense prediction and
  DG. In remote sensing, Earth-Adapter~\cite{EarthAdapter}
introduces a frequency-aware PEFT design for geospatial domain gaps, and
CrossEarth-Gate~\cite{CrossEarthGate} explores adaptive module selection for
efficient cross-domain remote sensing semantic segmentation. Despite these
advances, how   explicit geometric priors
from auxiliary modalities such as DSM can steer VFM adaptation  remains
insufficiently studied. Existing works emphasize  general remote sensing
pretraining,  single-modal adaptation   for cross-domain segmentation, or
multimodal adaptation for in-domain
segmentation\mbox{\cite{MFNet,MMDINO,MoBaNet}}\hskip0pt
,  while the role of
auxiliary modalities in improving generalization to unseen domains remains
underexplored.

\section{Proposed Method}
\label{sec:method}
In this section, we present the design of GeoDINO from the perspective of
its central idea: DSM-derived geometric structure, when injected as an
explicit prior into the self-attention of a frozen VFM, can guide the
optical encoder toward geometry-consistent
representations without full fine-tuning. Section~\ref{sec:network} first
presents the overall end-to-end pipeline, including dual-stream encoding,
geometry-aware adaptation, cross-modal fusion, UNetFormer decoding, and the
multitask training flow. Within this framework,
Section~\ref{sec:geo_prior} presents GeoPrior for extracting the structural
signal from DSM, Section~\ref{sec:geo_lora} describes GeoLoRA for injecting
it during parameter-efficient adaptation, Section~\ref{sec:fusion}
introduces DeSCA for selective cross-modal fusion,
and Section~\ref{sec:training} presents the mmMIM objective for training
under incomplete observations.

To make the subsequent formulation easier to follow, Table~\ref{tab:method_notation}
summarizes the main mathematical variables. The notation
is organized according to the GeoDINO pipeline: multimodal inputs, geometry-prior
construction, feature extraction and fusion, and the mmMIM training
objective.

\begin{table}[t]
\centering
\scriptsize
\caption{Definitions of key variables used in the GeoDINO method}
\label{tab:method_notation}
\fontsize{5.8}{6.8}\selectfont
\setlength{\tabcolsep}{2.0pt}
\renewcommand{\arraystretch}{1.0}
\begin{tabularx}{\columnwidth}{@{}l>{\raggedright\arraybackslash}Xl@{}}
\toprule
Symbol & Definition & Module \\
\midrule
\rowcolor{gray!25}
\multicolumn{3}{l}{$\blacktriangledown$ Multimodal inputs and dense prediction}\\
$\mathbf{I}$ & Optical image in $\mathbb{R}^{3\times H\times W}$. & Input \\
$\mathbf{M}$ & DSM map in $\mathbb{R}^{1\times H\times W}$, replicated when fed to the VFM encoder. & Input \\
$H,W,C$ & Image height, image width, and number of semantic classes. & Segmentation \\
$\hat{\mathbf{Y}},\mathbf{Y}$ & Ground-truth semantic map and predicted dense logits. & Decoder \\
\midrule
\rowcolor{gray!25}
\multicolumn{3}{l}{$\blacktriangledown$ DSM-derived geometry prior}\\
$\mathbf{D}^{s},\mathbf{D}^{d},\mathbf{D}^{r}$ & Pairwise spatial, elevation, and roughness distance matrices. & GeoPrior \\
$\alpha_s,\alpha_d,\alpha_r$ & Learnable weights for fusing the three geometric distance views. & GeoPrior \\
$\tau$ & Temperature parameter controlling the strength of the geometry bias. & GeoPrior \\
$\mathbf{G},\boldsymbol{\Gamma}$ & Additive log-space geometry bias and its multiplicative prior kernel. & \makecell[l]{GeoPrior/\\GeoLoRA} \\
\midrule
\rowcolor{gray!25}
\multicolumn{3}{l}{$\blacktriangledown$ Adaptation, fusion, and training objective}\\
$K$ & Number of selected transformer blocks where GeoLoRA injects $\boldsymbol{\Gamma}$. & GeoLoRA \\
$\mathbf{F}^{opt}_{\xi},\mathbf{F}^{d}_{\xi}$ & Multi-scale optical and DSM feature maps at pyramid stage $\xi\in\{1,2,3,4\}$. & \makecell[l]{Feature\\Pyramid} \\
$\mathbf{F}_{\xi}$ & DeSCA-fused multimodal feature map at stage $\xi$. & DeSCA \\
$\mathbf{m},\rho,s,u$ & Block-wise mask, masking ratio, block size, and sampled modality gate. & mmMIM \\
\makecell[l]{$\mathcal{L}_{seg},\mathcal{L}_{mim},$\\$\lambda_{mim}$} & Segmentation loss, reconstruction loss, and auxiliary loss weight. & Training \\
\bottomrule
\end{tabularx}
\end{table}

\subsection{Overview of GeoDINO}
\label{sec:network}

\begin{figure*}[t]
  \centering
  \includegraphics[width=0.90\textwidth]{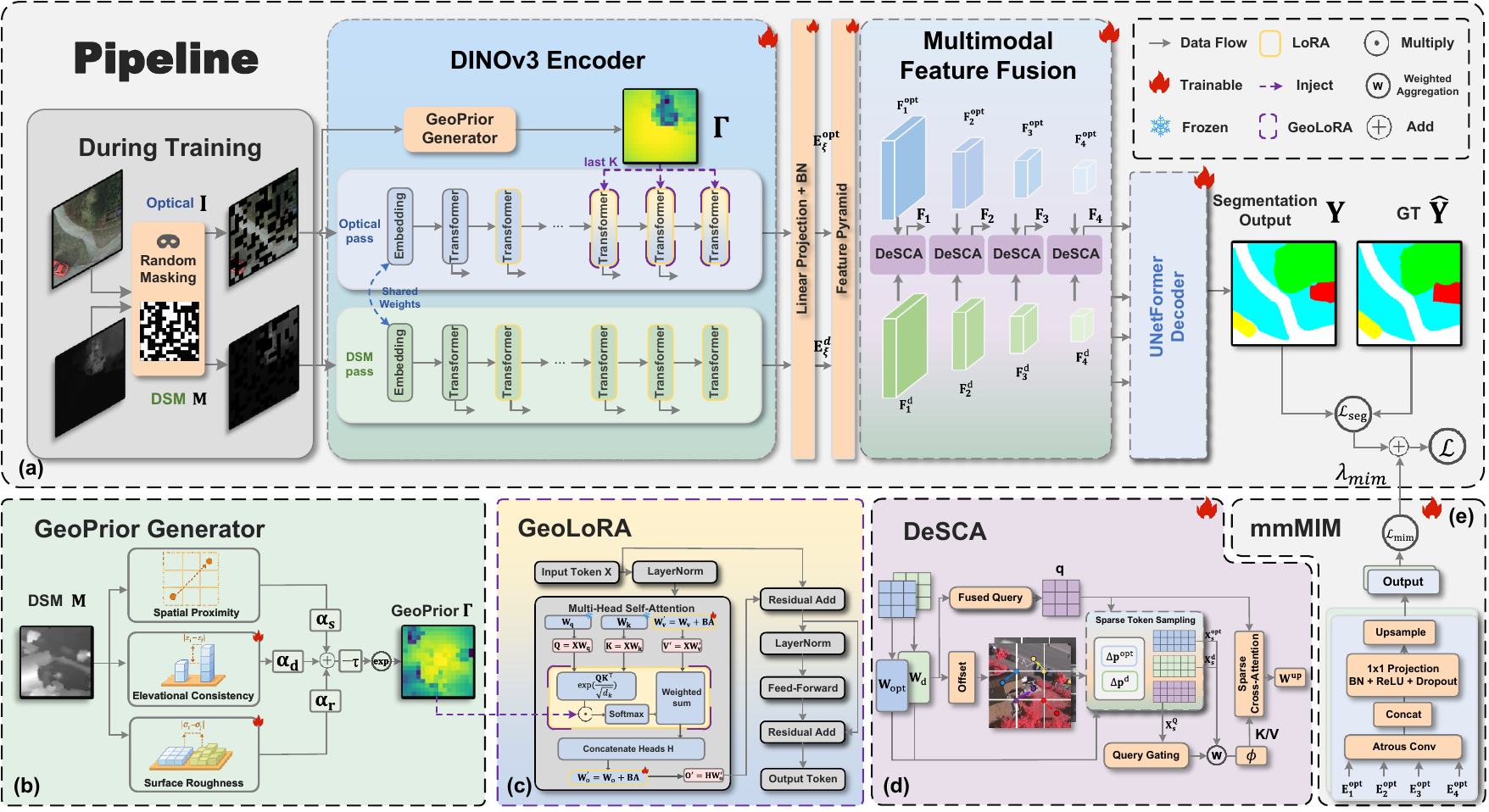}
  \caption{
  Overview of the proposed GeoDINO framework.
  (a) Overall training pipeline. A random single-modality mask is applied during training, and the masked optical or DSM input is fed into a dual-stream DINOv3 encoder. The extracted multilevel features are projected into a feature pyramid, fused by DeSCA, and decoded by UNetFormer for segmentation. The total objective combines the segmentation loss and the mmMIM loss.
  (b) GeoPrior Generator. A geometry prior $\boldsymbol{\Gamma}$ is derived from the DSM by combining spatial proximity, elevation consistency, and surface roughness cues.
  (c) GeoLoRA. The DSM-derived geometry prior is injected into selected self-attention layers of the optical encoder as a geometry-aware bias, while LoRA performs parameter-efficient adaptation.
  (d) DeSCA. The proposed DeSCA fusion module constructs fused queries, predicts modality-specific deformable offsets, performs sparse token sampling, and applies query gating for optical--DSM feature fusion.
  (e) mmMIM. A multimodal MIM branch reconstructs the masked modality from  the  multilevel  prefusion  features  of its own stream  and provides an auxiliary regularization signal through $\mathcal{L}_{mim}$.
  }

  \label{fig:overall_framework}
\end{figure*}

As shown in Fig.~\ref{fig:overall_framework}(a), the proposed GeoDINO follows a
dual-stream encoding--fusion--decoding pipeline. It takes an optical image
$\mathbf{I}\in\mathbb{R}^{3\times H\times W}$ and a DSM map
$\mathbf{M}\in\mathbb{R}^{1\times H\times W}$ as inputs, where $H$ and $W$ represent
the height and width of the inputs, respectively. For compatibility with the
foundation encoder, the DSM is  normalized,  replicated to three channels,  and
processed with the
same input transform as the optical imagery.  Replication
only provides a compatible tensor shape and does not make the DSM statistically
similar to the pretraining images; the DSM pass is adapted with its own
low-rank parameters, as described below.

We adopt a shared DINOv3 foundation encoder as the backbone to encode the two
modalities via two forward passes with   modality-conditioned adapter
routing. The pretrained encoder weights are frozen and shared, whereas each
adapted linear layer selects the low-rank parameters of the active modality.
The geometry bias is enabled only in the optical pass: GeoPrior derives an
explicit prior $\boldsymbol{\Gamma}$ from the current DSM input, and GeoLoRA
injects $\boldsymbol{\Gamma}$ as an attention bias into the last $K$
transformer blocks. From four
selected transformer layers, we extract multilevel tokens, project them
to a unified channel dimension, and reshape them into 2-D feature maps
$\mathbf{E}^{opt}_{\xi}$ and $\mathbf{E}^{d}_{\xi}$. Because
a plain ViT preserves a single $H/16\times W/16$ token grid at every
depth, we further resample these maps with the feature pyramid
module to four hierarchical scales of size
$(H/2^{\xi +1})\times(W/2^{\xi+1})$, producing multiscale optical and DSM feature maps
$\mathbf{F}^{opt}_{\xi}$ and $\mathbf{F}^{d}_{\xi}$, where $\xi\in\{1,2,3,4\}$.

The two pyramids are fused stage-by-stage via DeSCA to obtain fused
features $\mathbf{F}_{\xi}$ ($\xi\in\{1,2,3,4\}$). DeSCA performs deformable
sparse sampling and point-wise modality dominance weighting to select which
cross-modal correspondences are fused, establishing long-range intermodal interactions via
sparse cross-attention. Finally, a UNetFormer decoder~\cite{UNetFormer} progressively recovers
spatial details and outputs dense logits
$\mathbf{Y}\in\mathbb{R}^{H\times W\times C}$ for $C$ classes.

During training, we further employ the mmMIM objective: at each
iteration, a block-wise mask is sampled, and only one modality (optical or DSM)
is randomly masked. Lightweight reconstruction heads then regress the
masked modality, computing the reconstruction loss exclusively on the
masked regions. The overall training procedure is summarized in
Algorithm~\ref{alg:multitask_training}.

\begin{algorithm}[t]
\footnotesize
\makeatletter\def\@fs@cfont{\footnotesize\bfseries}\makeatother
\caption{\footnotesize Multitask Training of GeoLoRA With Multimodal MIM}
\label{alg:multitask_training}
\begin{algorithmic}[1]
\Require optical image $\mathbf{I}$, DSM map $\mathbf{M}$, segmentation label $\hat{\mathbf{Y}}$;
max iterations $T$; masking ratio $\rho$, block size $s$; mask generator $p_m(\rho,s)$;
DeSCA fusion $f_{\text{fuse}}$; decoder $f_{\text{dec}}$.
\For{$t = 1$ \textbf{to} $T$}
  \State Generate block-wise mask $\mathbf{m}$ with $p_m(\rho,s)$.
  \State Sample modality gate $u \sim \mathrm{Bernoulli}(0.5)$.
  \If{$u = 0$}
    \State Generate masked optical $\tilde{\mathbf{I}}$ by (\ref{eq:mmmim_mask_opt}).
    \State Compute $\boldsymbol{\Gamma}$ from unmasked DSM by (\ref{eq:gamma_def}).
  \Else
    \State Generate masked DSM $\tilde{\mathbf{M}}$ by (\ref{eq:mmmim_mask_dsm}).
    \State Compute $\boldsymbol{\Gamma}$ from masked DSM by (\ref{eq:gamma_def}).
  \EndIf
  \State GeoLoRA encoders produce maps $\mathbf{E}^{opt}_{\xi}$ and $\mathbf{E}^{d}_{\xi}$.
  \State The feature pyramid gives $\mathbf{F}^{opt}_{\xi}$ and $\mathbf{F}^{d}_{\xi}$.
  \For{$\xi=1$ \textbf{to} $4$}
    \State $\mathbf{F}_{\xi} \gets f_{\text{fuse}}(\mathbf{F}^{opt}_{\xi}, \mathbf{F}^{d}_{\xi})$
  \EndFor
  \State Predict segmentation $\mathbf{Y} \gets f_{\text{dec}}(\mathbf{F}_{1},\mathbf{F}_{2},\mathbf{F}_{3},\mathbf{F}_{4})$.
  \State Compute segmentation loss $\mathcal{L}_{seg}$ with $(\mathbf{Y}, \hat{\mathbf{Y}})$.
  \If{$u = 0$}
    \State Reconstruct $\hat{\mathbf{I}}$ by (\ref{eq:mmmim_recon}).
    \State Compute $\mathcal{L}_{mim}$ by (\ref{eq:mmmim_loss}).
  \Else
    \State Reconstruct $\hat{\mathbf{M}}$ by (\ref{eq:mmmim_recon}).
    \State Compute $\mathcal{L}_{mim}$ by (\ref{eq:mmmim_loss}).
  \EndIf
  \State Compute total loss $\mathcal{L}$ by (\ref{eq:overall_loss}).
  \State Update trainable parameters via back-propagation.
\EndFor
\end{algorithmic}
\end{algorithm}

\subsection{Geometry Prior Generator}
\label{sec:geo_prior}

  The DSM helps to separate categories with similar optical
appearance\mbox{\cite{Jin2022Fusion,vFuseNet}}, and it has been found to improve
cross-city accuracy on the ISPRS data\mbox{\cite{Petrich3D,Langenkamp2026}}.
To guide the adaptation with a signal that remains comparable across
domains, we first extract an explicit
structural prior from  the relative geometry of  the DSM  rather than from its absolute
heights. Inspired by DFormerv2~\cite{DFormerv2}, we
introduce a geometry prior generator tailored to remote sensing imagery.
Given a DSM map $\mathbf{M}\in\mathbb{R}^{H\times W}$, the generator learns
to produce a data-driven structural constraint for attention guidance. The
overall architecture of this module is illustrated in
Fig.~\ref{fig:overall_framework}(b).

\subsubsection{Multiview Geometric Feature Extraction}
Inside the generator, we quantify the geometric affinity between
any pair of patches $i$ and $j$ across three complementary dimensions.
This multiview approach ensures that the model captures both
macroscopic structure and microscopic texture:

\paragraph{Spatial proximity ($\mathbf{D}^s$)}
To enforce local inductive bias, we measure the physical distance
between patches based on their grid coordinates $(u, v)$.
We measure spatial proximity between patches using the Manhattan distance
($L_1$ norm), which matches the patch-grid structure of vision transformers:
\begin{equation}
D_{ij}^{s} = \frac{|u_i - u_j| + |v_i - v_j|}{C_s}.
\end{equation}

\paragraph{Elevational consistency ($\mathbf{D}^d$)}
To capture the height difference between two patches, we extract a
representative height value $z_i$ for each patch from the DSM $\mathbf{M}$.
Specifically, we use adaptive max pooling to obtain the patch-level height  on
the $16\times16$ token grid of a $256\times256$ input, which preserves
prominent local elevation responses and sharp height discontinuities between
neighboring patches
\begin{equation}
z_i = \text{MaxPool}(\mathbf{M})_i, \quad
D_{ij}^{d} = \frac{|z_i - z_j|}{C_d}.
\end{equation}

\paragraph{Surface roughness ($\mathbf{D}^r$)}
We introduce a textural metric to resolve the ambiguity between objects with
similar heights, such as flat roofs and tree canopies. For each position
$\delta$,  the local roughness $\sigma_{\delta}$   is computed as the standard deviation within the
corresponding nonoverlapping $16\times16$-px patch, from the mean and second raw moment
\begin{equation}
\sigma_{\delta} =
\sqrt{\text{Avg}(\mathbf{M}^2)_\delta - \left(\text{Avg}(\mathbf{M})_\delta\right)^2},
\quad
D_{ij}^{r} = \frac{|\sigma_i - \sigma_j|}{C_r}.
\end{equation}

The constants $C_s$, $C_d$, and $C_r$ denote the maximum values used to
normalize the corresponding distance matrices to the range $[0,1]$.  $C_s$ is
fixed by the token grid, whereas $C_d$ and $C_r$ are taken over the current
mini-batch in every forward pass, during both training and inference.

\subsubsection{Adaptive Fusion and Prior Definition}
To synthesize these metrics into a unified constraint,
the generator employs a learnable fusion mechanism.
For notational clarity, we describe the computation for a single attention
head; the same operation is applied independently
to each head. For each geometric component $m \in \{s, d, r\}$, we define
a learnable scalar parameter $\alpha_{m}$, kept positive by a softplus
reparameterization.
The module aggregates these
weighted distances and scales them by a   learnable  temperature parameter $\tau$
\begin{equation}
\mathbf{G} = -\tau \cdot \left(
\alpha_{s} \mathbf{D}^{s} +
\alpha_{d} \mathbf{D}^{d} +
\alpha_{r} \mathbf{D}^{r}
\right)
\end{equation}
where $\mathbf{G}$ represents the additive geometry bias
in the log-space. Finally, we formally define the
geometry prior matrix $\boldsymbol{\Gamma}$ as the multiplicative kernel
derived from this bias
\begin{equation}
\label{eq:gamma_def}
\boldsymbol{\Gamma} \triangleq \exp\!\big(\mathbf{G}\big)
\end{equation}
where $\exp(\cdot)$ denotes the element-wise exponential function. The resulting geometry prior $\boldsymbol{\Gamma}$ suppresses
connections between geometrically incoherent regions and is used in the
decoupled adaptation strategy described in
Section~\ref{sec:geo_lora}.
\subsection{Geometry-Aware Low-Rank Adaptation (GeoLoRA)}
\label{sec:geo_lora}

Given the geometry prior, the next step is to use it in the adaptation of the
frozen VFM. PEFT is used for this adaptation because it gives accurate
in-domain multimodal segmentation\mbox{\cite{MFNet}} and generalizes better
than full fine-tuning\mbox{\cite{Rein}}. In GeoDINO, PEFT adds low-rank
updates to the frozen weights. Learned from the single source domain alone,
these updates can fit the appearance of the source domain, which differs in an
unseen domain. To constrain these updates, we propose \mbox{GeoLoRA}, a
framework that couples the geometry prior with PEFT. The overall design of
GeoLoRA is illustrated in
Fig.~\ref{fig:overall_framework}(c). GeoLoRA wraps the transformer layers
and injects the prior $\boldsymbol{\Gamma}$ directly into the attention mechanism.

\subsubsection{Geometry-Aware Attention Injection}
We first define how the geometry prior is integrated into the
transformer blocks. The pretrained foundation model does not
explicitly encode the elevation-based boundaries of remote
sensing data. We therefore implement a geometry-aware wrapper
around the self-attention layers.
During the forward pass, we modulate the semantic affinity
using the geometry prior $\boldsymbol{\Gamma}$ derived in (\ref{eq:gamma_def}).
We first compute the attention score matrix $\mathbf{S}=[S_{ij}]$ with
$S_{ij}=(\langle \mathbf{q}_i, \mathbf{k}_j \rangle)/(\sqrt{d_k})$.
The rectified geometry-aware attention matrix
$\hat{\mathbf{A}}=[\hat{A}_{ij}]$ is then computed as
\begin{equation}
\label{eq:geo_attention_final}
\tilde{\mathbf{S}} = \exp\!\big(\mathbf{S}\big) \odot \boldsymbol{\Gamma}, \qquad
\hat{A}_{ij} =
\frac{\tilde{S}_{ij}}{\sum_{j'} \tilde{S}_{ij'}},
\end{equation}
where $\odot$ denotes element-wise multiplication.
When the semantic query-key matching produces a high score
for a geometrically invalid pair such as a roof and a nearby tree,
a near-zero value in $\Gamma_{ij}$ will suppress this connection.
Crucially, this explicit structural guidance relieves the
query and key projections from the burden of relearning
geometry from scratch. In implementation, this multiplication is performed
efficiently as an additive bias in the
log-probability space before the Softmax operation.

\subsubsection{Selective Semantic Adaptation}
Motivated by the geometry injection mechanism described above,
we adopt a decoupled adaptation strategy.
We freeze the query and key projections
($\mathbf{W}_{q}, \mathbf{W}_{k}$) during training and leave the
adaptation to the value, output, and feed-forward layers. This decouples
structural guidance from semantic content adaptation. We apply LoRA~\cite{LoRA} exclusively
to the value ($\mathbf{W}_v$), output ($\mathbf{W}_o$) projections,
and the feed-forward network (FFN) layers.
For a frozen weight $\mathbf{W}_0$, the update is parameterized
by low-rank matrices $\mathbf{B}$ (initialized to zero) and
$\mathbf{A}$ (Gaussian initialization)
\begin{equation}
\mathbf{W}' = \mathbf{W}_0 + \mathbf{B}\mathbf{A}, \quad \text{where } \mathbf{W}_0 \in \{\mathbf{W}_v, \mathbf{W}_o, \mathbf{W}_{\text{ffn}}\}.
\end{equation}
Consequently, given input features $\mathbf{X}$, the forward process uses
frozen   query/key projection weights and adapted value projections
\begin{equation}
\mathbf{Q} = \mathbf{X}\mathbf{W}_{q}, \quad
\mathbf{K} = \mathbf{X}\mathbf{W}_{k}, \quad
\mathbf{V}' = \mathbf{X}\mathbf{W}'_v.
\end{equation}
The geometry-rectified attention weights first aggregate the adapted values,
and the adapted output projection then produces the attention-block output
\begin{equation}
\mathbf{H} = \hat{\mathbf{A}}\mathbf{V}', \quad
\mathbf{O}' = \mathbf{H}\mathbf{W}'_o.
\end{equation}
In this design,  $\boldsymbol{\Gamma}$   adds a DSM-derived bias to
the query--key matching, and the LoRA-adapted $\mathbf{V}'$, $\mathbf{O}'$, and FFN layers
  learn  task-relevant   features.

\subsubsection{Selective Deep-Layer Injection}
We apply this geometry-aware modulation only in the last $K$ ViT layers,
leaving the early layers unchanged. This preserves general-purpose
representations in the shallow stages and imposes structural constraints
in the deeper semantic layers.

\subsection{Cross-Modal Fusion Design}
\label{sec:fusion}

Fusing the two modality streams
after geometry-aware encoding requires
  selecting which cross-modal interactions are fused. This is needed because,
in an unseen domain, the relation between the optical image and the DSM is not
always the same as in the source domain, for example, when the DSM is
misregistered or noisy. Cross-modal attention lets each
stream draw on the complementary cues of the other and improves in-domain
accuracy\mbox{\cite{FTransUNet,CMX}}. Global cross-modal attention, a common choice in existing multimodal frameworks, incurs quadratic
complexity and attends to all cross-modal token pairs. Inspired by deformable attention~\cite{DAT} and the dynamic
sparse fusion strategy~\cite{KTB}, we introduce DeSCA, a \emph{2-D}
deformable sparse fusion operator{} that attends to a sparse set of salient
tokens to capture long-range intermodal dependencies. The workflow of the proposed DeSCA module is illustrated in
Fig.~\ref{fig:overall_framework}(d).

At the $\xi$th pyramid level, DeSCA takes the optical and DSM feature maps
$\mathbf{F}^{opt}_{\xi}, \mathbf{F}^{d}_{\xi}\in
\mathbb{R}^{C_{\xi}\times H_{\xi}\times W_{\xi}}$ as inputs. We flatten them into token
sequences $\mathbf{X}^{opt}_{\xi}, \mathbf{X}^{d}_{\xi}\in
\mathbb{R}^{N_{\xi}\times C_{\xi}}$, where $N_{\xi}=H_{\xi}W_{\xi}$. DeSCA performs cross-modal
fusion through four components: low-rank query
construction, modality-specific deformable point prediction, sparse token
sampling with modality dominance weighting, and sparse cross-attention
aggregation.

\paragraph{Low-rank query construction}
To reduce computational cost, we first project both modalities into a low-rank
space using linear projections, yielding $\tilde{\mathbf{X}}^{opt}_{\xi}
= \mathbf{X}^{opt}_{\xi}\mathbf{W}^{down}_{opt}$ and
$\tilde{\mathbf{X}}^{d}_{\xi}
= \mathbf{X}^{d}_{\xi}\mathbf{W}^{down}_{d}$.
Here, $\mathbf{W}^{down}_{opt}, \mathbf{W}^{down}_{d}\in \mathbb{R}^{C_{\xi}\times \tilde{C}}$
are modality-specific linear down-projection matrices and $\tilde{C}\ll C_{\xi}$.
We then reshape the
projected tokens into 2-D feature maps
$\tilde{\mathbf{F}}^{opt}_{\xi},\tilde{\mathbf{F}}^{d}_{\xi}\in
\mathbb{R}^{\tilde{C}\times H_{\xi}\times W_{\xi}}$,
and construct a coarse fused query carrier
\begin{equation}
\mathbf{Q}=\psi\!\left([\tilde{\mathbf{F}}^{opt}_{\xi};\tilde{\mathbf{F}}^{d}_{\xi}]\right),
\qquad
\mathbf{q}=\varphi_q(\mathbf{Q}).
\end{equation}
$\psi$ can be instantiated as Conv-BN-ReLU and $\varphi_q$ is a $1\times1$ projection.

\paragraph{Modality-Specific Deformable Point Prediction}
To avoid dense all-to-all matching, DeSCA predicts modality-specific offsets on a
downsampled grid with stride $r$.
Let $\mathbf{p}\in\mathbb{R}^{H_{\xi,r}\times W_{\xi,r}\times 2}$ be the uniform reference grid
in normalized coordinates, where $H_{\xi,r}=(H_{\xi})/r$ and $W_{\xi,r}=(W_{\xi})/r$.
We predict offsets and obtain deformed points for each modality
\begin{align}
\Delta \mathbf{p}^{opt}
&= \varphi_{off}(\tilde{\mathbf{F}}^{opt}_{\xi}),
&
\Delta \mathbf{p}^{d}
&= \varphi_{off}(\tilde{\mathbf{F}}^{d}_{\xi}),
\label{eq:fusion_offsets}\\
\mathbf{p}^{opt}
&= \text{clip}(\mathbf{p}+\Delta \mathbf{p}^{opt}),
&
\mathbf{p}^{d}
&= \text{clip}(\mathbf{p}+\Delta \mathbf{p}^{d}).
\nonumber
\end{align}
Here, $\varphi_{off}(\cdot)$ is a lightweight offset predictor (depth-wise conv with stride $r$
followed by normalization, GELU, and $1\times1$ conv), and $\text{clip}(\cdot)$ clamps
coordinates into $[-1,1]$.

\paragraph{Sparse token sampling and modality dominance weighting}
We use the \emph{union} of the two deformed point sets to sample salient tokens from both modalities.
Denote $\mathcal{S}(\cdot,\cdot)$ as grid sampling.
Let $N_{\xi,s} = H_{\xi,r}W_{\xi,r}$. We construct sparse tokens by sampling each modality at both point sets
\begin{align}
\mathbf{X}^{opt}_{s}
&=
\Big[
\mathcal{S}(\tilde{\mathbf{F}}^{opt}_{\xi},\mathbf{p}^{opt});
\ \mathcal{S}(\tilde{\mathbf{F}}^{opt}_{\xi},\mathbf{p}^{d})
\Big]
\label{eq:fusion_sample_opt}\\
\mathbf{X}^{d}_{s}
&=
\Big[
\mathcal{S}(\tilde{\mathbf{F}}^{d}_{\xi},\mathbf{p}^{opt});
\ \mathcal{S}(\tilde{\mathbf{F}}^{d}_{\xi},\mathbf{p}^{d})
\Big].
\label{eq:fusion_sample_dsm}
\end{align}
$\mathbf{X}^{opt}_{s}$ and $\mathbf{X}^{d}_{s}$ are sparse token sets with shape
$\tilde{C}\times(2N_{\xi,s})$.

To weight the two modalities at each sampled point, we further compute point-wise modality dominance weights
using the same sampled locations as in (\ref{eq:fusion_sample_opt}) and (\ref{eq:fusion_sample_dsm})
\begin{align}
\mathbf{X}^{Q}_{s}
&=
\Big[
\mathcal{S}(\mathbf{q},\mathbf{p}^{opt});
\ \mathcal{S}(\mathbf{q},\mathbf{p}^{d})
\Big]
\label{eq:fusion_sample_q}\\
[\mathbf{w}^{opt};\mathbf{w}^{d}]
&=
\text{Softmax}\!\big(\omega(\mathbf{X}^{Q}_{s})\big)
\label{eq:fusion_modality_weight}\\
\mathbf{X}_{w}
&=
\mathbf{w}^{opt}\odot \mathbf{X}^{opt}_{s}
\;+\;
\mathbf{w}^{d}\odot \mathbf{X}^{d}_{s}.
\label{eq:fusion_weighted_tokens}
\end{align}
$\omega(\cdot)$ is a lightweight gating network (two $1\times1$ convolutions with ReLU) that outputs
two modality logits per sampled point; the Softmax is applied along the modality dimension to produce
$\mathbf{w}^{opt},\mathbf{w}^{d}\in\mathbb{R}^{(2N_{\xi,s})\times 1}$ with $w^{opt}_j+w^{d}_j=1$.

\paragraph{Sparse cross-attention aggregation}
The dominance-weighted sparse tokens are projected into keys and values
as $\mathbf{K}=\varphi_k(\mathbf{X}_{w})$ and
$\mathbf{V}=\varphi_v(\mathbf{X}_{w})$, and dense queries attend only to
these sparse candidates
\begin{equation}
\mathbf{A}
=
\text{Softmax}\!\left(
\frac{\mathbf{q}^\top\mathbf{K}}{\sqrt{d}}
+\mathbf{B}_{\text{rpe}}
\right).
\label{eq:fusion_attn}
\end{equation}
The fused output is then obtained by
$\mathbf{O}=\varphi_o(\mathbf{A}\mathbf{V})$.
Here, $\varphi_k$, $\varphi_v$, and $\varphi_o$ denote the key, value, and output
projection layers, respectively. $\mathbf{q}$ is reshaped into $N_{\xi}$ dense query vectors, while
$\mathbf{K}$ and $\mathbf{V}$ contain $2N_{\xi,s}$ sparse sampled candidates;
$d$ denotes the per-head channel dimension, and
$\mathbf{B}_{\text{rpe}}$ is a learnable relative positional bias over
query--sampled-point pairs.

Finally, we mix the deformable fusion output with an identity path and restore
the original channel width through
\begin{equation}
\mathbf{F}
=
\lambda_{def}\odot \mathbf{O}
+
\lambda_{id}\odot \mathbf{Q},
\qquad
\mathbf{F}_{\xi}
=
\text{reshape}(\mathbf{F})\mathbf{W}^{up}.
\label{eq:fusion_resmix}
\end{equation}
Here, $\lambda_{def}$ and $\lambda_{id}$ are learnable channel-wise mixing coefficients, and
$\mathbf{W}^{up}$ is a linear up-projection.

Overall, this 2-D formulation is well suited to dense prediction: it reduces the complexity from
$\mathcal{O}(N_{\xi}^2)$ to $\mathcal{O}(N_{\xi}\cdot N_{\xi,s})$, while the deformable offsets let each
modality select its own sampling locations.

\subsection{Multitask Training Pipeline}
\label{sec:training}

When both modalities are always complete during training, the model can come
to rely on their co-occurrence in the source domain, which does not always hold
in an unseen domain. Masking one modality counters this reliance: the masked
regions have to be segmented from the other modality and reconstructed from
the visible context. We
therefore incorporate an mmMIM auxiliary task inspired by MultiMAE~\cite{MultiMAE}; a masked
reconstruction objective on source data has also been used for domain-generalized remote
sensing segmentation\mbox{\cite{CrossEarth}}. The overall objective jointly
optimizes supervised segmentation on fused optical--DSM features and
MIM-based reconstruction of randomly masked observations. Rather than
masking both modalities simultaneously, we adopt random single-modality
block masking so that the   segmentation loss is always computed with one
intact stream while the reconstruction term regularizes the corrupted stream.
The mmMIM branch is illustrated in
Fig.~\ref{fig:overall_framework}(e).

\subsubsection{Multimodal Masked Image Modeling}
Given an optical image $\mathbf{I}\in\mathbb{R}^{3\times H\times W}$ and a DSM map
$\mathbf{M}\in\mathbb{R}^{1\times H\times W}$, we generate a binary mask
$\mathbf{m}\in\{0,1\}^{H\times W}$ at block granularity.
Let $p$ be the backbone patch size and $s=p\cdot s_b$ be the masking block size,
where $s_b$ is a scaling factor.
We uniformly sample a subset of blocks according to a masking ratio $\rho$ and
upsample the resulting block mask to image resolution using nearest-neighbor
interpolation, where $\mathbf{m}(i,j)=1$ indicates a masked pixel.

Unlike standard single-modality MIM, we perform random single-modality masking:
  in each training iteration, we randomly choose one modality and mask it in all
samples of the mini-batch
\begin{align}
\tilde{\mathbf{I}} &= \mathbf{I}\odot(1-\mathbf{m}),\quad \tilde{\mathbf{M}}=\mathbf{M}
\label{eq:mmmim_mask_opt}\\
\text{or}\qquad
\tilde{\mathbf{I}} &= \mathbf{I},\quad \tilde{\mathbf{M}}=\mathbf{M}\odot(1-\mathbf{m})
\label{eq:mmmim_mask_dsm}
\end{align}
where $\odot$ denotes element-wise multiplication with broadcasting across channels.
  When the optical image is masked, the intact DSM still conditions the
optical features through the geometry bias of GeoLoRA; when the DSM is masked,
the geometry bias is computed from the masked DSM.

We attach lightweight reconstruction heads for each modality and predict the
masked content from the corresponding modality stream features.
Using the feature maps of the corresponding modality stream before the feature
pyramid, we reconstruct
\begin{equation}
\hat{\mathbf{I}}
=
g_{opt}\!\left(\{\mathbf{E}^{opt}_{\xi}\}_{\xi=1}^{4}\right),
\qquad
\hat{\mathbf{M}}
=
g_{dsm}\!\left(\{\mathbf{E}^{d}_{\xi}\}_{\xi=1}^{4}\right).
\label{eq:mmmim_recon}
\end{equation}
Here, $g_{opt}$ and $g_{dsm}$ denote the modality-specific reconstruction
heads, and the subscripts index the four selected transformer layers from
shallow to deep.

\subsubsection{Overall Objective}
The mmMIM loss is computed only on masked pixels to avoid trivial identity learning
\begin{equation}
\scalebox{0.92}{$\displaystyle
\mathcal{L}_{mim}^{opt}
=
\frac{\|(\hat{\mathbf{I}}-\mathbf{I})\odot \mathbf{m}\|_{1}}{\|\mathbf{m}\|_{1}+\epsilon},
\quad
\mathcal{L}_{mim}^{d}
=
\frac{\|(\hat{\mathbf{M}}-\mathbf{M})\odot \mathbf{m}\|_{1}}{\|\mathbf{m}\|_{1}+\epsilon}
$}
\label{eq:mmmim_loss}
\end{equation}
where $\epsilon$ is a small constant for numerical stability. For each sample,
we activate only the loss corresponding to the masked modality in
(\ref{eq:mmmim_mask_opt}) and (\ref{eq:mmmim_mask_dsm}) and denote the active
term by $\mathcal{L}_{mim}$.

The final multitask objective is
\begin{equation}
\mathcal{L}
=
\mathcal{L}_{seg}
+
\lambda_{mim}\,\mathcal{L}_{mim}
\label{eq:overall_loss}
\end{equation}
where $\mathcal{L}_{seg}$ is the supervised segmentation loss on the decoder output and
$\lambda_{mim}$ balances the auxiliary mmMIM regularization.
During inference, the reconstruction heads are discarded, and only the segmentation
branch is used.
\section{Experiments and Discussion}

\subsection{Datasets and Benchmark Protocol}
\label{sec:dataset_protocol}

We conduct experiments on three multimodal remote sensing benchmarks:
ISPRS Vaihingen, ISPRS Potsdam, and MMHunan. All three benchmarks are
used for both in-domain evaluation and DG assessment.

\subsubsection{Datasets}
\label{sec:datasets}

\paragraph{ISPRS Vaihingen}
The ISPRS Vaihingen benchmark comprises 33 very high-resolution true
orthophoto (TOP) tiles with varying spatial sizes, averaging
$2500\times2000$ pixels. Each tile provides three spectral channels
(near-infrared, red, and green) together with an aligned   DSM, all at a ground sampling distance (GSD) of 9\,cm.
The task involves six semantic classes: Impervious Surface (Surf),
Buildings (Bldg), Low Vegetation (Vegt), Trees (Tree), Cars (Car),
and Clutter. We adopt the official split by selecting tiles 1, 3, 7,
11, 13, 17, 23, 26, 28, 32, 34, and 37 for training, while tiles 5,
15, 21, and 30 are reserved for testing. All original tiles are
cropped into $256\times256$ patches.

\paragraph{ISPRS Potsdam}
The ISPRS Potsdam benchmark consists of 38 high-resolution TOP tiles,
each of size $6000\times6000$ pixels. It provides multiple spectral
configurations, including RGB,   IRRG (near-infrared, red, and green), and four-channel RGB-IR, at a
GSD of 5\,cm, along with aligned DSMs. The task shares the same six
categories (Surf, Bldg, Vegt, Tree, Car, and Clutter) as Vaihingen.
We follow a custom split, using tiles 6\_10, 7\_10, 2\_12, 3\_11,
2\_10, 7\_8, 5\_10, 3\_12, 5\_12, 7\_11, 7\_9, 6\_9, 7\_7, 4\_12,
6\_8, 6\_12, 6\_7, 4\_11 for training, and reserving tiles 4\_10,
5\_11, 2\_11, 3\_10, 6\_11, 7\_12 for testing. Similar to Vaihingen,
all tiles are cropped into $256\times256$ patches.

\paragraph{MMHunan}
The MMHunan dataset \cite{DKDFN} exhibits a distinct spatial granularity compared
to the ISPRS benchmarks, featuring a GSD of approximately 10\,m.
Collected in Hunan, China, it contains 500 Sentinel-2 image patches,
each of size $256\times256$ pixels, paired with co-registered DSMs.
We select the red, green, and blue bands to form the visible imagery.
The dataset includes seven land-cover classes: cropland, forest, grassland, wetland, water, unused land, and built-up area. For in-domain
segmentation, the 400 official training patches and the 50 official
validation patches are used for training, and the 50 official test patches are used for testing. For regional transfer, the 500 patches are divided at the
median longitude of their centers into a western and an eastern part of 250
patches each; within each part, the official training patches are used for
training and the official validation and test patches for testing.

\begin{table*}[t]
\centering
\scriptsize
\caption{Comparison of methods on ISPRS Vaihingen and Potsdam under in-domain and joint region--spectral transfer settings. For each task, class-wise accuracy, OA, and mIoU are reported. The best results are in bold and the second best results are underlined (\%).  Subscripts indicate the publication year of each method}
\label{tab:ss_6tasks}
\setlength{\tabcolsep}{3.0pt}
\renewcommand{\arraystretch}{1.02}
  \resizebox{0.97\textwidth}{!}{
\begin{tabular}{l l c r r r r r r r c r r r r r r r}
\toprule
\multirow{2}{*}[-1.1ex]{Method} &
\multirow{2}{*}[-1.1ex]{Backbone} &
\multicolumn{1}{c}{Setting} &
\multicolumn{6}{c}{Classes --- Acc} &
\multirow{2}{*}[-1.1ex]{mIoU} &
\multicolumn{1}{c}{Setting} &
\multicolumn{6}{c}{Classes --- Acc} &
\multirow{2}{*}[-1.1ex]{mIoU} \\
\cmidrule(lr){3-3}\cmidrule(lr){4-9}\cmidrule(lr){11-11}\cmidrule(lr){12-17}
& & Train $\rightarrow$ Test &
Surf & Bldg & Vegt & Tree & Car & Total &
& Train $\rightarrow$ Test &
Surf & Bldg & Vegt & Tree & Car & Total & \\
\midrule
FuseNet\pubyear{2016}\cite{FuseNet} & VGG16 &
\multirow{15}{*}{\parbox[c]{18mm}{\centering V $\rightarrow$ V\\(In-domain)}} &
92.07 & 97.15 & 79.40 & 91.52 & 92.62 & 91.38 & 81.89 &
\multirow{15}{*}{\parbox[c]{18mm}{\centering V $\rightarrow$ P(r)\\(Unseen Region and Spectral Band)}} &
62.12 & 82.55 & 44.23 & 3.56 & 31.76 & 52.29 & 31.70 \\
PSPNet\pubyear{2017}\cite{PSPNet} & ResNet101 & &
92.75 & 96.47 & 81.31 & 89.39 & 88.66 & 91.13 & 81.87 &
& 57.49 & 75.80 & 9.38 & 3.92 & 72.32 & 42.76 & 31.00 \\
TransUNet\pubyear{2021}\cite{TransUNet} & R50-ViT-B & &
93.09 & 96.04 & 77.76 & \textbf{93.62} & 91.48 & 91.57 & 82.83 &
& 73.74 & 88.72 & 48.52 & 6.52 & \underline{84.68} & 59.51 & 42.96 \\
FTUNetFormer\pubyear{2022}\cite{UNetFormer} & Swin-B & &
92.81 & 97.26 & 78.86 & 92.02 & 90.98 & 91.64 & 82.97 &
& 72.75 & 88.93 & 57.84 & 4.35 & 66.26 & 60.65 & 43.00 \\
CMGFNet\pubyear{2022}\cite{CMGFNet} & ResNet34 & &
92.62 & 96.92 & 81.05 & 91.79 & 91.01 & 91.79 & 83.18 &
& 79.14 & 78.62 & 41.33 & 5.73 & 51.31 & 56.50 & 38.57 \\
CMFNet\pubyear{2022}\cite{CMFNet} & VGG16 & &
90.74 & 96.49 & \underline{83.68} & 91.02 & 89.05 & 91.37 & 82.93 &
& 64.51 & 88.43 & 2.22 & 0.05 & 72.41 & 45.92 & 20.93 \\
ESANet\pubyear{2021}\cite{ESANet} & ResNet34 & &
91.41 & 96.47 & 77.98 & 90.39 & 80.98 & 90.38 & 79.42 &
& 70.65 & 87.83 & 23.04 & 1.11 & 15.94 & 51.32 & 28.54 \\
FTransUNet\pubyear{2024}\cite{FTransUNet} & R50-ViT-B & &
93.36 & \textbf{98.64} & 80.39 & 92.08 & 90.18 & 92.47 & 83.97 &
& 78.02 & 87.51 & 65.84 & \underline{34.71} & 40.67 & 67.71 & 48.26 \\
FTransDeepLab\pubyear{2025}\cite{FTransDeepLab} & SegFormer-B2 & &
93.52 & \underline{98.38} & 80.77 & 91.94 & 89.79 & 92.47 & 84.13 &
& 85.68 & 92.48 & 70.05 & 7.37 & 78.52 & 68.72 & 51.54 \\
SiMultiF\pubyear{2025}\cite{SiMultiF} & Swin-T & &
92.46 & 97.86 & 78.69 & 92.35 & 89.81 & 91.77 & 83.03 &
& 79.27 & 85.70 & 55.45 & 6.95 & 65.06 & 61.63 & 45.71 \\
EG-CMFNet\pubyear{2026}\cite{EGCMFNet} & BiFormer-T & &
93.37 & 97.48 & 82.97 & 89.66 & 92.01 & 91.96 & 83.58 &
& 77.10 & 92.92 & 31.58 & 10.90 & 71.93 & 58.71 & 42.79 \\
MMLoRA\pubyear{2025}\cite{MFNet} & SAM-L & &
\textbf{93.87} & 98.18 & 81.46 & \underline{92.72} & 91.57 & \textbf{92.83} & 84.93 &
& \underline{87.32} & 92.29 & \underline{77.62} & 14.82 & 76.43 & 71.83 & 56.69 \\
MMAdapter\pubyear{2025}\cite{MFNet} & SAM-L & &
\underline{93.71} & 97.75 & 82.94 & 91.90 & 91.02 & \underline{92.68} & \underline{85.16} &
& 73.86 & 93.80 & \textbf{80.98} & 29.85 & 78.62 & 71.17 & 57.66 \\
MM-DINO\pubyear{2026}\cite{MMDINO} & DINOv3-L & &
93.08 & 95.42 & 81.50 & 92.11 & \underline{93.95} & 91.63 & 84.23 &
& 84.38 & \textbf{98.78} & 72.08 & 29.86 & \textbf{90.53} & \underline{74.13} & \underline{60.32} \\
\rowcolor{gray!25}
\textbf{GeoDINO (Ours)} & DINOv3-L & &
93.58 & 98.26 & \textbf{83.82} & 91.33 & \textbf{94.13} & \textbf{92.83} & \textbf{85.92} &
& \textbf{93.97} & \underline{96.28} & 70.47 & \textbf{41.80} & 83.03 & \textbf{77.62} & \textbf{65.27} \\
\midrule
FuseNet\pubyear{2016}\cite{FuseNet} & VGG16 &
\multirow{15}{*}{\parbox[c]{18mm}{\centering P(r) $\rightarrow$ P(r)\\(In-domain)}} &
90.79 & 97.72 & 87.18 & 86.06 & 96.43 & 90.03 & 84.30 &
\multirow{15}{*}{\parbox[c]{18mm}{\centering P(r) $\rightarrow$ V\\(Unseen Region and Spectral Band)}} &
6.29 & 98.94 & 0.84 & 47.18 & 8.13 & 43.99 & 18.14 \\
PSPNet\pubyear{2017}\cite{PSPNet} & ResNet101 & &
92.72 & 97.77 & 88.66 & 86.23 & 96.37 & 90.95 & 85.77 &
& 23.99 & 78.19 & 25.62 & 65.49 & 53.28 & 51.47 & 37.33 \\
TransUNet\pubyear{2021}\cite{TransUNet} & R50-ViT-B & &
93.46 & 97.87 & 87.70 & 88.13 & \underline{96.92} & 91.13 & 86.19 &
& 57.57 & 85.17 & 17.16 & 47.67 & 42.16 & 56.83 & 42.58 \\
FTUNetFormer\pubyear{2022}\cite{UNetFormer} & Swin-B & &
93.12 & 97.54 & 87.52 & 88.79 & 96.20 & 91.10 & 85.93 &
& 59.09 & 90.15 & 10.71 & 73.16 & 34.89 & 63.98 & 46.21 \\
CMGFNet\pubyear{2022}\cite{CMGFNet} & ResNet34 & &
92.86 & 97.65 & 87.70 & 86.60 & 96.62 & 90.52 & 85.17 &
& 13.40 & 82.81 & 2.17 & 53.22 & 21.45 & 42.83 & 23.42 \\
CMFNet\pubyear{2022}\cite{CMFNet} & VGG16 & &
92.46 & 97.90 & 87.68 & 87.60 & 95.63 & 90.52 & 84.93 &
& 3.61 & 83.70 & 7.86 & 76.03 & 0.21 & 46.87   & 21.39 \\
ESANet\pubyear{2021}\cite{ESANet} & ResNet34 & &
91.69 & 97.48 & 85.94 & 82.61 & 95.57 & 89.42 & 82.98 &
& 56.06 & 70.95 & 3.10 & 0.20 & 11.38 & 37.69 & 18.03 \\
FTransUNet\pubyear{2024}\cite{FTransUNet} & R50-ViT-B & &
93.94 & 97.93 & 87.41 & 88.90 & 96.86 & 91.51 & 86.59 &
& 34.97 & 96.62 & 30.86 & 73.18 & 38.92 & 62.73 & 42.18 \\
FTransDeepLab\pubyear{2025}\cite{FTransDeepLab} & SegFormer-B2 & &
93.33 & 97.84 & 87.99 & 88.55 & 96.66 & 91.28 & 86.18 &
& 43.56 & 95.82 & 18.26 & 56.53 & 21.81 & 58.48 & 36.50 \\
SiMultiF\pubyear{2025}\cite{SiMultiF} & Swin-T & &
93.34 & 97.80 & 87.82 & 87.84 & 96.64 & 91.18 & 85.96 &
& 52.60 & 96.55 & 2.54 & 74.46 & 34.74 & 63.12 & 38.83 \\
EG-CMFNet\pubyear{2026}\cite{EGCMFNet} & BiFormer-T & &
93.28 & 98.04 & 87.91 & 87.25 & 96.70 & 91.28 & 86.11 &
& 4.10 & \textbf{99.61} & 6.12 & 36.10 & 42.50 & 42.00 & 22.32 \\
MMLoRA\pubyear{2025}\cite{MFNet} & SAM-L & &
93.89 & 98.24 & 87.91 & 88.49 & 96.72 & 91.69 & 86.95 &
& 68.94 & 57.81 & \textbf{94.46} & 20.59 & 82.13 & 55.38 & 53.07 \\
MMAdapter\pubyear{2025}\cite{MFNet} & SAM-L & &
93.45 & \underline{98.25} & 88.86 & \textbf{90.04} & 96.55 & \underline{91.88} & \underline{87.14} &
& 50.50 & 96.80 & 17.63 & 77.82 & 54.64 & 66.11 & 49.63 \\
MM-DINO\pubyear{2026}\cite{MMDINO} & DINOv3-L & &
\textbf{94.08} & 97.90 & \underline{88.90} & 88.52 & \underline{96.92} & 91.84 & 86.98 &
& \underline{77.56} & \underline{99.31} & 34.79 & \underline{84.68} & \textbf{90.13} & \underline{79.07} & \underline{65.53} \\
\rowcolor{gray!25}
\textbf{GeoDINO (Ours)} & DINOv3-L & &
\underline{93.97} & \textbf{98.48} & \textbf{89.80} & \underline{89.56} & \textbf{97.19} & \textbf{92.62} & \textbf{87.96} &
& \textbf{85.67} & 96.30 & \underline{42.65} & \textbf{89.75} & \underline{84.33} & \textbf{82.86} & \textbf{68.04} \\
\bottomrule
\end{tabular}
}

\end{table*}

\subsubsection{Benchmark Protocol}
\label{sec:protocol}

We evaluate GeoDINO under two test conditions: matched training and test
distributions (in-domain) and unseen target domains (DG).
For notation, we denote Vaihingen as V, Potsdam with RGB input as P(r),
Potsdam with IRRG input as P(i), and MMHunan as H, with H(W) and H(E) for
its western and eastern parts.

For DG evaluation, we consider three domain shifts:
unseen region, unseen spectral band, and their combination.
Unseen-region transfer includes P(i)$\rightarrow$V, V$\rightarrow$P(i),
H(W)$\rightarrow$H(E), and H(E)$\rightarrow$H(W);
unseen-spectral transfer includes P(i)$\rightarrow$P(r) and
P(r)$\rightarrow$P(i); and the joint region--spectral shift includes
P(r)$\rightarrow$V and V$\rightarrow$P(r).
In every transfer setting, a model is trained on one labeled source domain only,
without any target-domain images, and is evaluated on the target domain without
adaptation.
We organize the experiments into two groups as follows.

\textbf{Experiment I} benchmarks GeoDINO against existing
state-of-the-art in-domain segmentation methods, with results in
Tables~\ref{tab:ss_6tasks} and~\ref{tab:mmhunan}.
To ensure strict comparability, all methods in this group are trained
with the same segmentation loss and evaluated on
V$\rightarrow$V, P(r)$\rightarrow$P(r), H$\rightarrow$H,
H(W)$\rightarrow$H(W), and H(E)$\rightarrow$H(E).
We further test the methods trained in this way on
V$\rightarrow$P(r), P(r)$\rightarrow$V, H(W)$\rightarrow$H(E), and
H(E)$\rightarrow$H(W) as a source-only transfer check.

\textbf{Experiment II} focuses on DG and covers the eight transfer settings above, with
results in Table~\ref{tab:dg_6tasks}.
We compare GeoDINO against three categories of baselines under DG
evaluation: source-only DA architectures, single-modal RSDG models, and GeoDINO with
different PEFT variants.
For the third category, we retain the dual-stream multimodal structure
of GeoDINO, including the backbone, fusion module, decoder, and
training pipeline, and apply different PEFT designs to both the optical and DSM branches.

\begin{table*}[t]
\centering
\scriptsize
\caption{Quantitative comparison on MMHunan under the official split and under a regional split into a western part, H(W), and an eastern part, H(E). For each setting, class-wise accuracy, OA, mF1, and mIoU are reported. The best results are in bold. The second best results are underlined (\%). Subscripts indicate the publication year of each method}
\label{tab:mmhunan}
\setlength{\tabcolsep}{2.4pt}
\renewcommand{\arraystretch}{1.0}
\newsavebox{\hunanregbox}\sbox{\hunanregbox}{%
\begin{tabular}{l l c *{8}{r} r r c *{8}{r} r r}
\midrule
\multirow{2}{*}[-1.1ex]{Method} &
\multirow{2}{*}[-1.1ex]{Backbone} &
\multicolumn{1}{c}{Setting} & \multicolumn{8}{c}{Classes --- Acc} & \multirow{2}{*}[-1.1ex]{mF1} & \multirow{2}{*}[-1.1ex]{mIoU} &
\multicolumn{1}{c}{Setting} & \multicolumn{8}{c}{Classes --- Acc} & \multirow{2}{*}[-1.1ex]{mF1} & \multirow{2}{*}[-1.1ex]{mIoU} \\
\cmidrule(lr){3-3}\cmidrule(lr){4-11}\cmidrule(lr){14-14}\cmidrule(lr){15-22}
& & Train $\rightarrow$ Test & Crop & Forest & Grass & Wetl & Water & Unus & Built & Total &  &  &
Train $\rightarrow$ Test & Crop & Forest & Grass & Wetl & Water & Unus & Built & Total &  &  \\
\midrule
FuseNet\pubyear{2016}\cite{FuseNet} & VGG16 &
\multirow{15}{*}{\parbox[c]{18mm}{\centering H(W) $\rightarrow$ H(W)\\(In-domain)}} &
68.04 & 86.08 & 43.64 & 13.37 & \underline{82.41} & 19.12 & 70.21 & 74.87 & 55.34 & 42.03 &
\multirow{15}{*}{\parbox[c]{18mm}{\centering H(W) $\rightarrow$ H(E)\\(Unseen Region)}} &
66.43 & 80.34 & 14.39 & 1.10 & 78.03 & 7.35 & 73.00 & 69.83 & 46.43 & 35.77 \\
PSPNet\pubyear{2017}\cite{PSPNet} & ResNet101 & &
61.36 & 89.33 & 29.46 & 6.89 & 61.25 & 27.73 & 69.92 & 72.30 & 52.63 & 39.13 &
& 60.49 & 76.98 & 12.63 & 0.00 & 74.32 & 15.40 & 75.86 & 66.74 & 44.54 & 33.53 \\
TransUNet\pubyear{2021}\cite{TransUNet} & R50-ViT-B & &
66.17 & 91.18 & 27.06 & 7.46 & 65.62 & \textbf{44.07} & 75.10 & 74.92 & 56.36 & 42.95 &
& 62.44 & 84.52 & 16.16 & \underline{4.67} & 57.79 & 37.80 & 71.14 & 64.11 & 45.95 & 33.57 \\
FTUNetFormer\pubyear{2022}\cite{UNetFormer} & Swin-B & &
67.63 & 89.12 & 43.73 & 12.15 & 65.00 & 34.48 & 77.53 & 75.99 & 58.98 & 45.17 &
& 66.25 & \textbf{84.86} & 16.25 & 0.99 & 65.12 & 27.81 & 79.88 & 68.45 & 48.24 & 36.82 \\
CMGFNet\pubyear{2022}\cite{CMGFNet} & ResNet34 & &
65.86 & 83.17 & \underline{46.56} & 3.95 & 72.77 & 28.86 & 71.74 & 72.70 & 53.37 & 40.22 &
& 58.73 & 78.58 & 15.72 & \textbf{5.04} & 78.25 & 16.65 & 77.91 & 68.12 & 47.13 & 35.71 \\
CMFNet\pubyear{2022}\cite{CMFNet} & VGG16 & &
62.95 & 87.18 & 43.95 & 8.69 & 66.86 & 29.96 & 75.15 & 73.90 & 55.25 & 41.76 &
& 53.13 & 77.12 & \textbf{25.59} & 1.62 & 63.73 & 24.19 & 71.24 & 61.27 & 42.85 & 31.15 \\
ESANet\pubyear{2021}\cite{ESANet} & ResNet34 & &
64.32 & 88.04 & 35.79 & 5.23 & 63.15 & 29.81 & 78.32 & 73.79 & 53.61 & 40.24 &
& 67.40 & 81.56 & 5.32 & 0.98 & \underline{80.04} & 18.03 & 64.56 & 69.69 & 45.87 & 35.01 \\
FTransUNet\pubyear{2024}\cite{FTransUNet} & R50-ViT-B & &
65.48 & \underline{91.74} & 30.54 & 8.17 & 72.11 & 29.26 & 63.65 & 74.66 & 55.25 & 41.90 &
& 72.96 & 83.60 & 14.23 & 2.66 & 70.51 & 24.54 & 57.27 & 68.43 & 47.21 & 35.22 \\
FTransDeepLab\pubyear{2025}\cite{FTransDeepLab} & SegFormer-B2 & &
\textbf{72.40} & 89.16 & 41.95 & 8.28 & 79.24 & 41.14 & 78.19 & \underline{77.85} & \underline{60.80} & \underline{47.76} &
& 64.24 & 82.59 & 17.21 & 0.93 & 75.60 & 35.78 & 80.75 & 70.54 & 49.96 & 38.18 \\
SiMultiF\pubyear{2025}\cite{SiMultiF} & Swin-T & &
70.10 & 90.10 & 45.10 & 9.60 & 69.48 & \underline{43.18} & 74.91 & 77.25 & 60.68 & 47.00 &
& \underline{73.08} & \underline{84.76} & 12.00 & 2.99 & 71.72 & \textbf{43.48} & \underline{82.50} & \underline{73.00} & \underline{52.29} & \underline{40.57} \\
EG-CMFNet\pubyear{2026}\cite{EGCMFNet} & BiFormer-T & &
69.54 & \textbf{92.07} & 32.46 & \underline{13.67} & 69.64 & 41.59 & \underline{78.35} & 77.24 & 60.21 & 46.53 &
& 62.07 & 84.62 & 12.59 & 0.52 & 77.49 & 33.67 & 80.97 & 70.70 & 49.55 & 38.28 \\
MMLoRA\pubyear{2025}\cite{MFNet} & SAM-L & &
71.14 & 91.10 & 37.54 & 11.17 & 78.83 & 36.72 & 67.20 & 77.08 & 59.41 & 45.98 &
& \textbf{75.16} & 78.11 & 19.21 & 0.00 & 63.37 & 19.23 & 80.57 & 69.38 & 47.58 & 36.56 \\
MMAdapter\pubyear{2025}\cite{MFNet} & SAM-L & &
67.85 & 89.10 & 45.24 & 4.75 & 68.65 & 35.66 & 71.53 & 75.79 & 57.19 & 43.71 &
& 70.82 & 80.19 & \underline{21.30} & 0.53 & 70.54 & 36.37 & 75.91 & 70.07 & 50.24 & 37.84 \\
MM-DINO\pubyear{2026}\cite{MMDINO} & DINOv3-L & &
\underline{71.55} & 87.20 & 40.95 & 9.68 & 78.23 & 38.23 & 74.44 & 76.14 & 59.28 & 45.88 &
& 69.91 & 74.99 & 11.43 & 1.67 & 76.98 & \underline{38.34} & 80.53 & 71.08 & 50.47 & 39.11 \\
\rowcolor{gray!25}
\textbf{GeoDINO (Ours)} & DINOv3-L & &
64.88 & 89.71 & \textbf{48.24} & \textbf{20.23} & \textbf{83.19} & 40.54 & \textbf{82.56} & \textbf{78.06} & \textbf{63.60} & \textbf{49.72} &
& 64.45 & 81.92 & 20.27 & 3.62 & \textbf{85.66} & 37.44 & \textbf{83.67} & \textbf{73.86} & \textbf{54.11} & \textbf{42.02} \\
\midrule
FuseNet\pubyear{2016}\cite{FuseNet} & VGG16 &
\multirow{15}{*}{\parbox[c]{18mm}{\centering H(E) $\rightarrow$ H(E)\\(In-domain)}} &
81.03 & \underline{89.61} & 21.95 & 15.06 & 81.85 & 33.61 & 77.33 & 78.88 & 58.30 & 46.41 &
\multirow{15}{*}{\parbox[c]{18mm}{\centering H(E) $\rightarrow$ H(W)\\(Unseen Region)}} &
47.89 & \textbf{91.64} & 37.31 & 11.51 & 62.50 & 44.13 & 70.95 & 71.96 & 54.46 & 40.45 \\
PSPNet\pubyear{2017}\cite{PSPNet} & ResNet101 & &
76.97 & 87.70 & 19.74 & 9.63 & 85.16 & 38.26 & 78.60 & 78.21 & 57.26 & 45.44 &
& 55.27 & 87.91 & 25.20 & 2.57 & 62.24 & 29.89 & 67.23 & 69.70 & 49.64 & 36.32 \\
TransUNet\pubyear{2021}\cite{TransUNet} & R50-ViT-B & &
84.44 & 87.62 & 17.36 & 20.96 & 83.97 & 45.34 & 79.29 & 80.66 & 60.57 & 48.98 &
& 62.36 & 83.05 & 31.70 & 19.55 & 67.59 & \underline{46.16} & \underline{79.47} & 71.16 & 56.36 & 41.93 \\
FTUNetFormer\pubyear{2022}\cite{UNetFormer} & Swin-B & &
\underline{85.95} & 85.67 & 28.13 & 22.41 & 83.25 & 37.60 & 82.56 & 81.14 & 61.94 & 50.00 &
& 57.19 & 88.20 & 33.90 & 11.05 & 67.84 & 33.40 & 77.88 & 72.62 & 54.76 & 40.81 \\
CMGFNet\pubyear{2022}\cite{CMGFNet} & ResNet34 & &
78.83 & 82.17 & 24.40 & 23.33 & 80.68 & \underline{51.20} & \underline{83.42} & 77.79 & 59.05 & 46.84 &
& 50.22 & 85.41 & \underline{44.22} & 5.60 & 68.73 & 41.86 & 78.34 & 70.95 & 53.07 & 39.55 \\
CMFNet\pubyear{2022}\cite{CMFNet} & VGG16 & &
78.99 & 88.22 & 26.04 & 11.68 & 82.95 & 31.48 & 81.13 & 78.78 & 58.29 & 46.46 &
& 57.35 & 89.17 & 33.45 & 8.37 & 70.73 & 41.32 & 78.20 & 73.38 & 55.81 & 42.26 \\
ESANet\pubyear{2021}\cite{ESANet} & ResNet34 & &
75.23 & 83.33 & 26.79 & 14.62 & 84.88 & 38.66 & 78.82 & 77.01 & 57.80 & 45.69 &
& 52.20 & 88.07 & 36.32 & 3.60 & 58.41 & 38.90 & 74.63 & 70.82 & 51.52 & 37.95 \\
FTransUNet\pubyear{2024}\cite{FTransUNet} & R50-ViT-B & &
83.12 & 87.41 & 17.89 & 25.00 & 80.19 & 38.57 & 79.64 & 79.22 & 59.80 & 47.82 &
& 58.19 & 89.76 & 26.28 & 10.20 & 70.37 & 44.39 & 74.69 & 72.83 & 55.02 & 41.42 \\
FTransDeepLab\pubyear{2025}\cite{FTransDeepLab} & SegFormer-B2 & &
83.20 & 87.68 & 29.39 & 20.12 & 84.16 & 38.64 & 83.36 & 81.04 & 61.94 & 49.98 &
& 60.55 & 87.91 & 36.25 & 16.69 & 66.95 & 39.21 & \textbf{81.18} & 73.77 & 57.64 & 43.27 \\
SiMultiF\pubyear{2025}\cite{SiMultiF} & Swin-T & &
83.63 & 86.22 & \textbf{35.81} & 27.20 & \underline{86.41} & 43.33 & 81.68 & 81.64 & 63.52 & \underline{51.39} &
& 62.38 & 88.18 & 37.67 & 13.40 & 68.15 & \textbf{46.96} & 78.21 & 74.26 & 57.66 & 43.71 \\
EG-CMFNet\pubyear{2026}\cite{EGCMFNet} & BiFormer-T & &
81.76 & \textbf{90.71} & 17.80 & 27.50 & 85.99 & 32.51 & \textbf{84.98} & \underline{81.83} & 62.18 & 50.30 &
& \textbf{70.96} & 87.89 & 36.29 & \underline{22.20} & 69.08 & 35.72 & 73.30 & \underline{75.30} & \underline{59.65} & \underline{45.19} \\
MMLoRA\pubyear{2025}\cite{MFNet} & SAM-L & &
\textbf{87.00} & 83.05 & \underline{32.80} & \textbf{32.69} & 83.27 & 44.74 & 78.89 & 80.88 & \underline{63.60} & 51.04 &
& 65.67 & 83.74 & 41.99 & 4.92 & 73.50 & 42.51 & 74.86 & 72.98 & 56.11 & 42.83 \\
MMAdapter\pubyear{2025}\cite{MFNet} & SAM-L & &
84.73 & 85.57 & 30.97 & 27.12 & 82.41 & 42.29 & 79.74 & 80.34 & 62.11 & 49.87 &
& 61.85 & 84.86 & \textbf{44.68} & 10.50 & \underline{78.61} & 34.87 & 76.97 & 73.59 & 57.21 & 43.70 \\
MM-DINO\pubyear{2026}\cite{MMDINO} & DINOv3-L & &
83.94 & 87.92 & 21.66 & 21.84 & 81.91 & 39.59 & 82.28 & 80.45 & 60.81 & 48.94 &
& 59.12 & \underline{90.10} & 30.29 & 8.52 & 72.76 & 33.31 & 74.66 & 73.62 & 54.92 & 41.65 \\
\rowcolor{gray!25}
\textbf{GeoDINO (Ours)} & DINOv3-L & &
83.73 & 86.28 & 29.88 & \underline{30.16} & \textbf{87.56} & \textbf{51.56} & 83.37 & \textbf{82.34} & \textbf{64.99} & \textbf{52.65} &
& \underline{69.88} & 88.34 & 38.90 & \textbf{34.47} & \textbf{84.02} & 31.95 & 65.96 & \textbf{76.02} & \textbf{61.92} & \textbf{47.68} \\
\bottomrule
\end{tabular}
}%
  \resizebox{0.97\textwidth}{!}{%
\begin{tabular}{@{}c@{}}
\begin{tabularx}{\wd\hunanregbox}{l l c *{10}{>{\centering\arraybackslash}X}}
\toprule
\multirow{2}{*}[-1.1ex]{Method} &
\multirow{2}{*}[-1.1ex]{Backbone} &
\multicolumn{1}{c}{Setting} &
\multicolumn{8}{c}{Classes --- Acc} &
\multirow{2}{*}[-1.1ex]{mF1} &
\multirow{2}{*}[-1.1ex]{mIoU} \\
\cmidrule(lr){3-3}\cmidrule(lr){4-11}
& & Train $\rightarrow$ Test &
\makebox[0pt]{Cropland} & \makebox[0pt]{Forest} & \makebox[0pt]{Grassland} & \makebox[0pt]{Wetland} & \makebox[0pt]{Water} & \makebox[0pt]{Unused Land} & \makebox[0pt]{Built-up Area} & \makebox[0pt]{Total} & & \\
\midrule
FuseNet\pubyear{2016}\cite{FuseNet} & VGG16 & \multirow{15}{*}{\parbox[c]{18mm}{\centering H $\rightarrow$ H\\(In-domain)}}
& 79.37 & 89.76 & 37.62 & 19.52 & 82.96 & 36.44 & 75.09 & 80.39 & 61.06 & 48.44 \\
PSPNet\pubyear{2017}\cite{PSPNet} & ResNet101 & 
& 76.05 & 89.46 & 33.70 & 25.69 & 82.62 & 42.05 & 75.82 & 79.47 & 62.01 & 48.62 \\
TransUNet\pubyear{2021}\cite{TransUNet} & R50-ViT-B & 
& 79.48 & 88.71 & 30.41 & 36.31 & 83.27 & 43.95 & 78.09 & 80.43 & 62.87 & 49.98 \\
FTUNetFormer\pubyear{2022}\cite{UNetFormer} & Swin-B & 
& 78.83 & 89.45 & 36.17 & 26.24 & \underline{89.24} & 41.57 & \textbf{83.61} & \underline{82.37} & 64.88 & 52.10 \\
CMGFNet\pubyear{2022}\cite{CMGFNet} & ResNet34 & 
& 77.70 & 86.99 & 35.49 & 19.03 & \textbf{89.83} & 31.55 & 77.41 & 80.27 & 60.96 & 48.38 \\
CMFNet\pubyear{2022}\cite{CMFNet} & VGG16 & 
& 77.85 & \underline{91.02} & 35.78 & 28.83 & 81.58 & 38.07 & 81.69 & 81.10 & 63.56 & 50.38 \\
ESANet\pubyear{2021}\cite{ESANet} & ResNet34 & 
& 71.02 & \textbf{91.61} & 36.14 & 13.07 & 87.60 & 26.78 & 73.46 & 79.44 & 58.87 & 46.37 \\
FTransUNet\pubyear{2024}\cite{FTransUNet} & R50-ViT-B & 
& 79.90 & 88.54 & 34.49 & 15.78 & 83.42 & 43.46 & 79.68 & 80.57 & 61.76 & 49.28 \\
FTransDeepLab\pubyear{2025}\cite{FTransDeepLab} & SegFormer-B2 & 
& 78.72 & 90.65 & 40.29 & 21.58 & 84.85 & 44.04 & 79.40 & 81.66 & 64.15 & 51.24 \\
SiMultiF\pubyear{2025}\cite{SiMultiF} & Swin-T & 
& \underline{80.91} & 89.06 & 40.14 & 36.27 & 84.65 & 38.69 & 80.50 & 81.80 & 65.21 & 52.05 \\
EG-CMFNet\pubyear{2026}\cite{EGCMFNet} & BiFormer-T & 
& 76.97 & 89.19 & 38.54 & 35.22 & 89.16 & 39.98 & \underline{82.25} & 81.85 & 66.27 & 52.93 \\
MMLoRA\pubyear{2025}\cite{MFNet} & SAM-L & 
& \textbf{81.15} & 86.15 & \underline{47.61} & 39.77 & 84.64 & \textbf{48.45} & 78.12 & 80.94 & 65.80 & 52.40 \\
MMAdapter\pubyear{2025}\cite{MFNet} & SAM-L & 
& 79.00 & 90.01 & \textbf{48.07} & \underline{44.67} & 87.13 & 41.37 & 75.97 & 82.11 & \underline{67.46} & \underline{53.87} \\
MM-DINO\pubyear{2026}\cite{MMDINO} & DINOv3-L & 
& 80.05 & 89.47 & 34.60 & 39.15 & 83.69 & 40.38 & 78.60 & 81.17 & 64.39 & 51.18 \\
\rowcolor{gray!25}
\textbf{GeoDINO (Ours)} & DINOv3-L & 
& 80.55 & 90.87 & 46.49 & \textbf{46.99} & 87.78 & \underline{47.04} & 74.46 & \textbf{82.80} & \textbf{68.67} & \textbf{55.07} \\
\end{tabularx}
\\
\usebox{\hunanregbox}
\end{tabular}%
}

\end{table*}

\subsection{Implementation Details}
\paragraph{Experimental settings}
All experiments were conducted on a workstation with one RTX 4090.
The GPU provides 24 GB of memory for all reported experiments.
For fair comparison, we implement all models in PyTorch.
Optimization uses AdamW with $\beta_1=0.9$ and $\beta_2=0.999$.
Weight decay is set to 0.01 to mitigate overfitting.
We standardize training across networks with a batch size of 10.
All models are trained for 50 epochs under identical protocols.
The initial learning rate is $1\times10^{-4}$ for all methods.
We apply cosine annealing after a five-epoch linear warmup.

\paragraph{Data augmentation and evaluation}
To improve robustness, we apply consistent online augmentations during
training. We use random vertical flips and horizontal mirroring with
$p=0.5$. Augmentations are applied consistently across all available
modalities. Inputs are cropped into fixed $256\times256$ patches for
training. Segmentation performance is evaluated by  overall accuracy (OA), mean F1  score  (mF1),
and mIoU. For efficiency analysis, we further report  floating-point operations (FLOPs),
trainable parameters, peak GPU memory, and inference speed  in frames per second (FPS).

 \paragraph{Loss functions}
The supervised segmentation loss is defined as the standard pixel-wise
cross-entropy
\begin{equation}
\mathcal{L}_{\mathrm{CE}}
= -\frac{1}{N}\sum_{i=1}^{N}\sum_{c=1}^{C} y_{i,c}\log p_{i,c}
\end{equation}
where $N$ denotes the number of pixels, $C$ is the number of semantic
classes, $y_{i,c}\in\{0,1\}$ indicates whether pixel $i$ belongs to
class $c$, and $p_{i,c}$ is the predicted probability of class $c$ at
pixel $i$.
In (\ref{eq:overall_loss}), the supervised term is set to
$\mathcal{L}_{seg}=\mathcal{L}_{\mathrm{CE}}$.

The weight $\lambda_{mim}$ in (\ref{eq:overall_loss}) is treated as a
hyperparameter. Following the protocol in Section~\ref{sec:protocol}, it is set
to 0 in \textbf{Experiment I}, so GeoDINO and the in-domain comparison methods
use only $\mathcal{L}_{\mathrm{CE}}$, ensuring a strictly supervised and
architecture-focused comparison. In \textbf{Experiment II}, \mbox{$\lambda_{mim}>0$}
for GeoDINO and all same-framework PEFT variants, while the other models follow
the loss settings specified in their original papers.

\subsection{Performance Comparison}
\begin{figure*}[t]
  \centering
  \includegraphics[width=0.83\textwidth]{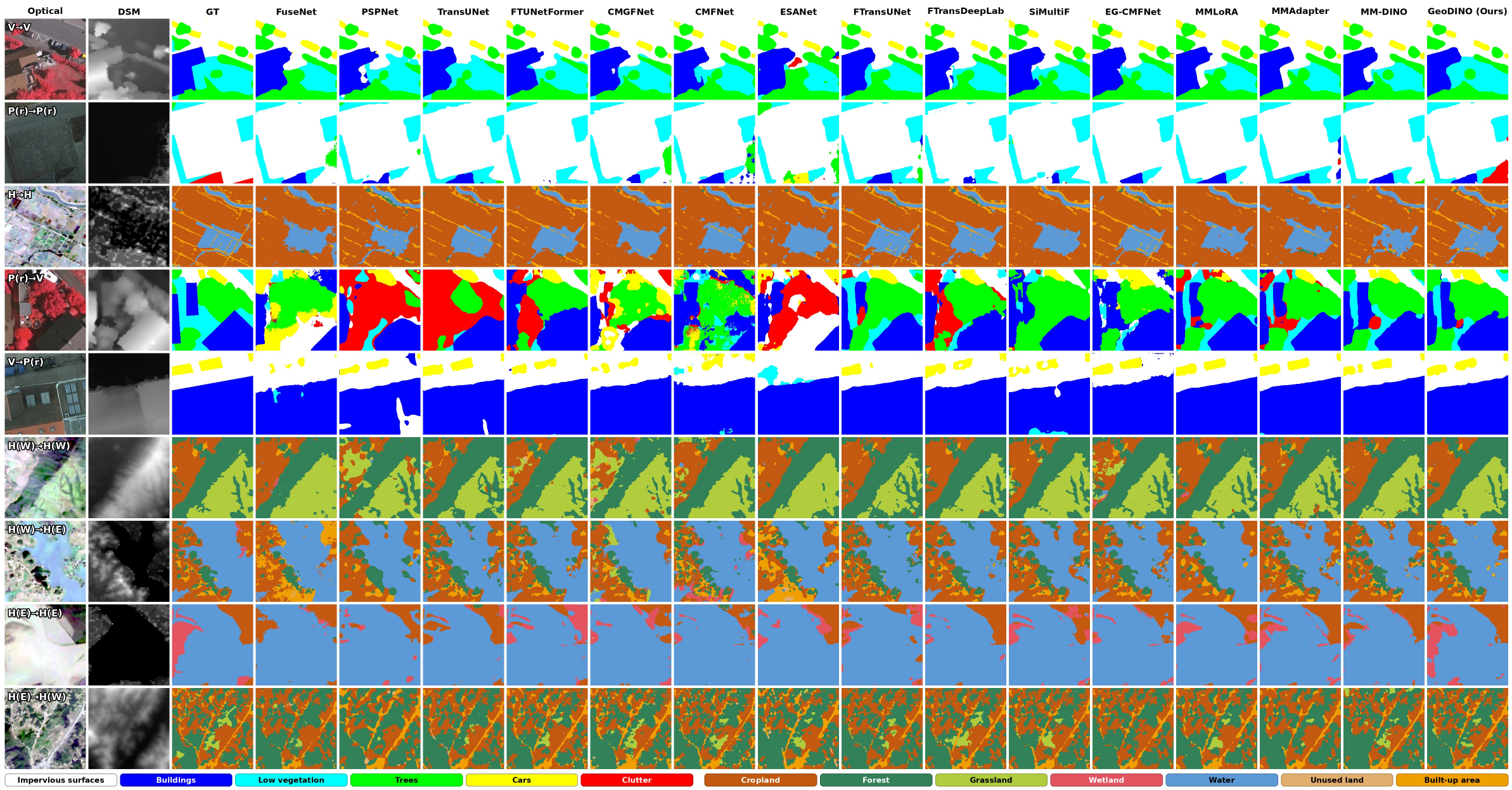}
  \caption{Visualizations of predicted segmentation maps on selected in-domain and   source-only  transfer tasks.}
  \label{fig:ss_vis}
\end{figure*}

\subsubsection{Experiment I (In-Domain Segmentation and Source-Only Transfer Check)}

Experiment~I compares GeoDINO with strong in-domain
segmentation methods, including FuseNet\cite{FuseNet}, PSPNet\cite{PSPNet}, TransUNet\cite{TransUNet},
FTUNetFormer\cite{UNetFormer}, CMGFNet\cite{CMGFNet}, CMFNet\cite{CMFNet}, ESANet\cite{ESANet}, FTransUNet\cite{FTransUNet},
FTransDeepLab\cite{FTransDeepLab}, SiMultiF\cite{SiMultiF}, EG-CMFNet\cite{EGCMFNet},
MMLoRA\cite{MFNet}, MMAdapter\cite{MFNet}, and MM-DINO\cite{MMDINO}.
The evaluation covers the standard in-domain settings
V$\rightarrow$V, P(r)$\rightarrow$P(r), and H$\rightarrow$H,
together with two   source-only transfer checks,
V$\rightarrow$P(r) and P(r)$\rightarrow$V, and the regional split of
MMHunan.
As reported in Tables~\ref{tab:ss_6tasks} and~\ref{tab:mmhunan}, GeoDINO delivers the most competitive
and the most consistent performance across all evaluated settings.

On Vaihingen, GeoDINO achieves 92.83\% OA and 85.92\% mIoU.
Its OA matches the best competing result achieved by MMLoRA,
while its mIoU exceeds MMAdapter by 0.76\%.
This establishes the best overall result on this benchmark.
At the class level, GeoDINO attains the highest accuracy on
low vegetation and cars, while remaining highly competitive on
the other categories.
Notably, the 94.13\% accuracy on cars surpasses all competing
methods, reflecting the
benefit of DSM-derived geometric cues for small-object
discrimination.

On Potsdam, GeoDINO again achieves the best overall performance,
with 92.62\% OA and 87.96\% mIoU.
Compared with the second-best MMAdapter, GeoDINO improves OA and mIoU by
0.74\% and 0.82\%, respectively.
It further obtains the best class-wise accuracy on
buildings, low vegetation, and cars,
showing clear advantages on both large structural regions
and small urban objects. The larger gains on Potsdam, which
has a finer GSD than Vaihingen (5 versus 9\,cm), suggest
that geometry-aware adaptation is more effective when
higher-resolution DSM inputs are available.

On MMHunan, despite the coarser spatial resolution and
larger scene variability, GeoDINO still achieves the best
overall results, reaching 82.80\% OA, 68.67\% mF1, and
55.07\% mIoU.
Compared with the strongest competing methods, it improves
OA by 0.43\% over FTUNetFormer and improves mF1 and mIoU
by 1.21\% and 1.20\%, respectively, over MMAdapter.
GeoDINO also achieves the best class-wise result on Wetland
and remains highly competitive on the remaining categories.
These consistent gains across three datasets with different
resolutions and land-cover distributions indicate broad   in-domain
applicability; cross-domain transfer is assessed in Experiment~II.

  The right half of Table~\ref{tab:ss_6tasks} applies the same
cross-entropy-trained models to the two joint region--spectral shifts,
 V$\rightarrow$P(r) and P(r)$\rightarrow$V.
Under V$\rightarrow$P(r), GeoDINO achieves 77.62\% OA and
65.27\% mIoU, outperforming the strongest competing method,
namely MM-DINO, by 3.49\% in OA and 4.95\% in mIoU.
Under P(r)$\rightarrow$V, GeoDINO reaches 82.86\% OA and
68.04\% mIoU, exceeding MM-DINO by 3.79\% in OA
and 2.51\% in mIoU.
These results reveal a clear contrast:
although existing segmentation baselines can remain competitive
under in-domain evaluation, their performance degrades
substantially under severe distribution shifts,
whereas GeoDINO preserves stronger transferability.
  The comparison with methods designed for domain shift is given in
Experiment~II. The attribution of the transfer gain to  geometry-aware
  adaptation is examined using controlled component and fusion ablations in
Section~\ref{sec:ablation}.

In the regional split of Table~\ref{tab:mmhunan}, GeoDINO leads in OA, mF1,
and mIoU within each part and on both transfers. Its in-domain mIoU is
49.72\% on H(W)$\rightarrow$H(W) and 52.65\% on H(E)$\rightarrow$H(E),
outperforming FTransDeepLab by +1.96\% and SiMultiF by +1.26\%, respectively.
Every method has lower mIoU on the other region. GeoDINO achieves 42.02\%
on H(W)$\rightarrow$H(E) and 47.68\% on H(E)$\rightarrow$H(W), outperforming
SiMultiF by +1.45\% and EG-CMFNet by +2.49\%, respectively.

Fig.~\ref{fig:ss_vis} presents qualitative comparisons of
segmentation maps on the ISPRS and MMHunan
tasks. On the ISPRS tiles, GeoDINO yields sharper and more accurate boundaries for
buildings, trees, and cars, leading to clearer separation
between adjacent regions.
It also preserves object integrity more effectively and
reduces fragmented or noisy predictions.
Overall, the visual results are cleaner and more
structurally consistent, which agrees well with the
quantitative improvements reported in
Tables~\ref{tab:ss_6tasks} and~\ref{tab:mmhunan}.

\subsubsection{Experiment II (Cross-Domain Comparison)}

\begin{table*}[t]
  \centering
  \scriptsize
  \caption{Comparison of DA methods, single-modal RSDG models, and GeoDINO with different PEFT variants under six cross-domain multimodal segmentation settings on the ISPRS benchmarks and two unseen-region settings on MMHunan. For each task, class-wise accuracy, OA, and mIoU are reported. The best results are in bold and the second best results are underlined (\%).  Subscripts indicate the publication year of each method}
  \label{tab:dg_6tasks}

  \setlength{\tabcolsep}{3.0pt}
  \renewcommand{\arraystretch}{1.02}
    \newsavebox{\dgispbox}\sbox{\dgispbox}{%
  \begin{tabular}{l c c r r r r r r r c r r r r r r r}
  \toprule
  \multirow{2}{*}[-1.1ex]{Method} &
  \multirow{2}{*}[-1.1ex]{Backbone} &
  \multicolumn{1}{c}{Setting} &
  \multicolumn{6}{c}{Classes --- Acc} &
  \multirow{2}{*}[-1.1ex]{mIoU} &
  \multicolumn{1}{c}{Setting} &
  \multicolumn{6}{c}{Classes --- Acc} &
  \multirow{2}{*}[-1.1ex]{mIoU} \\
  \cmidrule(lr){3-3}\cmidrule(lr){4-9}\cmidrule(lr){11-11}\cmidrule(lr){12-17}
  & & Train $\rightarrow$ Test &
  Surf & Bldg & Vegt & Tree & Car & Total &
  & Train $\rightarrow$ Test &
  Surf & Bldg & Vegt & Tree & Car & Total & \\
  \midrule
  DAFormer\pubyear{2022} (source-only)\cite{DAFormer} & MiT-B5 &
  \multirow{13}{*}{\parbox[c]{40mm}{\centering P(i) $\rightarrow$ V\\(Unseen Region)}} &
  60.05 & 93.46 & 41.71 & 91.89 & 45.27 & 75.09 & 54.58 &
  \multirow{13}{*}{\parbox[c]{40mm}{\centering V $\rightarrow$ P(i)\\(Unseen Region)}} &
  77.20 & 85.62 & 85.00 & 16.88 & 70.07 & 68.67 & 52.09 \\
  HRDA\pubyear{2022} (source-only)\cite{HRDA} & MiT-B5 & & 67.76 & 90.38 & 43.24 & 87.51 & 63.52 & 75.55 & 59.16 & & 81.88 & 62.55 & \textbf{93.10} & 23.72 & 74.27 & 66.40 & 52.32 \\
  GeoSA-BaSA\pubyear{2025}\cite{GeoSABaSA} & DINOv2-L & & 79.00 & 92.18 & \underline{71.62} & 85.04 & 83.00 & 82.69 & 71.45 & & 81.72 & 88.76 & 84.17 & 39.68 & 81.30 & 74.38 & 62.11 \\
  CrossEarth\pubyear{2025}\cite{CrossEarth} & DINOv2-L & & 83.10 & 90.88 & 70.84 & 78.68 & 78.61 & 82.29 & 71.63 & & 77.09 & \textbf{98.53} & 56.20 & 35.49 & 88.27 & 69.32 & 56.94 \\

  Frozen & DINOv3-L & & 51.83 & 97.98 & 40.30 & 88.16 & 82.95 & 73.36 & 58.13 & & 87.09 & \underline{97.15} & 82.46 & 46.25 & 87.91 & 79.34 & 66.53 \\
  Full Fine-Tune & DINOv3-L & & 54.63 & 97.55 & 31.40 & \underline{92.98} & 53.68 & 73.48 & 53.25 & & 90.59 & 96.28 & 90.37 & 44.23 & 83.63 & \underline{81.94} & \underline{70.03} \\
  Adapter\pubyear{2019}\cite{Adapter} & DINOv3-L & & 79.09 & \underline{99.28} & 62.16 & 85.63 & \underline{84.52} & 84.14 & 70.52 & & \underline{92.02} & 95.15 & 83.16 & 43.09 & 89.62 & 79.68 & 68.35 \\
  LoRA\pubyear{2022}\cite{LoRA} & DINOv3-L & & 80.46 & 95.37 & \textbf{84.90} & 77.66 & 75.71 & 85.03 & \underline{71.95} & & 91.18 & 95.43 & 91.66 & 39.09 & 91.83 & 80.70 & 69.55 \\
  VPT\pubyear{2022}\cite{VPT} & DINOv3-L & & 67.77 & 98.62 & 64.40 & 86.80 & 79.47 & 81.50 & 67.28 & & 83.85 & 88.63 & 89.77 & 34.08 & 71.66 & 75.12 & 60.81 \\
  Rein\pubyear{2024}\cite{Rein} & DINOv3-L & & \underline{83.18} & 98.84 & 41.01 & \textbf{93.26} & 74.23 & 83.46 & 68.34 & & 90.94 & 94.56 & 81.81 & 43.56 & 83.41 & 78.87 & 66.65 \\
  Earth-Adapter\pubyear{2026}\cite{EarthAdapter} & DINOv3-L & & \textbf{84.44} & 98.56 & 54.69 & 92.18 & 75.51 & \underline{85.70} & 71.56 & & 90.92 & 94.22 & 73.49 & \underline{52.24} & \underline{92.30} & 78.44 & 67.70 \\
  MM-DINO\pubyear{2026}\cite{MMDINO} & DINOv3-L & & 81.34 & \textbf{99.32} & 63.09 & 84.01 & \textbf{94.60} & 84.63 & 71.72 & & 80.72 & 97.02 & \underline{92.15} & 50.34 & \textbf{95.80} & 79.80 & 69.01 \\
  \rowcolor{gray!25}
  \textbf{GeoDINO (Ours)} & DINOv3-L & & 79.78 & 98.47 & 67.95 & 89.82 & 81.30 & \textbf{86.08} & \textbf{73.64} & & \textbf{93.56} & 92.05 & 83.71 & \textbf{63.60} & 83.32 & \textbf{82.36} & \textbf{71.85} \\

  \midrule
  DAFormer\pubyear{2022} (source-only)\cite{DAFormer} & MiT-B5 &
  \multirow{13}{*}{\parbox[c]{40mm}{\centering P(r) $\rightarrow$ P(i)\\(Unseen Spectral Band)}} &
  91.45 & 97.86 & 74.55 & 83.77 & 95.27 & 87.29 & 81.01 &
  \multirow{13}{*}{\parbox[c]{40mm}{\centering P(i) $\rightarrow$ P(r)\\(Unseen Spectral Band)}} &
  92.30 & 98.27 & 86.19 & 80.00 & 97.11 & 89.66 & 83.78 \\
  HRDA\pubyear{2022} (source-only)\cite{HRDA} & MiT-B5 & & 91.08 & 96.21 & 72.97 & 90.20 & \textbf{98.01} & 87.30 & 80.54 & & 94.13 & 97.78 & 80.62 & 79.68 & \underline{97.45} & 89.02 & 82.80 \\
  GeoSA-BaSA\pubyear{2025}\cite{GeoSABaSA} & DINOv2-L & & 91.11 & 97.46 & 80.03 & 87.80 & 95.65 & 89.19 & 82.88 & & 90.56 & 97.31 & 81.02 & 86.81 & 96.40 & 89.29 & 83.66 \\
  CrossEarth\pubyear{2025}\cite{CrossEarth} & DINOv2-L & & 86.60 & 96.56 & 59.92 & \textbf{92.60} & 93.39 & 84.52 & 78.61 & & 93.01 & 96.97 & \textbf{90.87} & 78.64 & 93.54 & 90.26 & 84.04 \\

  Frozen & DINOv3-L & & 92.29 & \underline{98.45} & 76.44 & 89.56 & 96.32 & 89.15 & 83.68 & & 92.74 & 97.99 & 85.66 & 83.29 & 97.14 & 90.53 & 85.17 \\
  Full Fine-Tune & DINOv3-L & & 91.51 & \textbf{98.75} & \textbf{84.86} & 89.54 & \underline{97.11} & \underline{90.57} & \underline{85.35} & & 94.74 & \textbf{98.73} & 85.80 & \underline{88.12} & 97.19 & 91.37 & 86.22 \\
  Adapter\pubyear{2019}\cite{Adapter} & DINOv3-L & & 89.68 & 97.96 & \underline{84.13} & \underline{92.29} & 92.29 & 90.37 & 85.00 & & 92.94 & 97.91 & 89.05 & 83.18 & \textbf{97.79} & 91.26 & 86.05 \\
  LoRA\pubyear{2022}\cite{LoRA} & DINOv3-L & & 92.44 & 98.38 & 80.27 & 91.27 & 97.08 & 90.29 & 84.89 & & 94.43 & 98.11 & 87.73 & 84.05 & 96.76 & 91.37 & 86.20 \\
  VPT\pubyear{2022}\cite{VPT} & DINOv3-L & & 91.95 & 98.17 & 82.74 & 90.32 & 96.41 & 90.38 & 85.08 & & 93.24 & 98.36 & \underline{89.92} & 84.18 & 97.10 & 91.48 & 86.33 \\
  Rein\pubyear{2024}\cite{Rein} & DINOv3-L & & 92.00 & 98.32 & 82.27 & 90.39 & 96.31 & 90.51 & 85.29 & & 93.11 & 98.36 & 88.16 & 87.26 & 96.85 & 91.64 & 86.53 \\
  Earth-Adapter\pubyear{2026}\cite{EarthAdapter} & DINOv3-L & & 90.10 & 97.69 & 84.10 & 91.21 & 96.39 & 90.39 & 85.19 & & \underline{94.75} & 98.09 & 86.44 & \textbf{89.37} & 96.87 & \underline{91.68} & \underline{86.79} \\
  MM-DINO\pubyear{2026}\cite{MMDINO} & DINOv3-L & & \textbf{93.39} & 98.09 & 78.90 & 92.15 & \underline{97.11} & 90.24 & 84.89 & & \textbf{94.81} & 97.89 & 87.16 & 81.57 & 97.17 & 90.88 & 85.45 \\
  \rowcolor{gray!25}
  \textbf{GeoDINO (Ours)} & DINOv3-L & & \underline{93.16} & 98.23 & 82.64 & 89.85 & 96.82 & \textbf{90.70} & \textbf{85.47} & & 93.15 & \underline{98.40} & 89.89 & 86.32 & 97.05 & \textbf{91.95} & \textbf{86.96} \\

  \midrule
  DAFormer\pubyear{2022} (source-only)\cite{DAFormer} & MiT-B5 &
  \multirow{13}{*}{\parbox[c]{40mm}{\centering P(r) $\rightarrow$ V\\(Unseen Region and Spectral Band)}} &
  51.99 & 69.87 & 19.52 & 87.21 & 58.97 & 60.97 & 49.48 &
  \multirow{13}{*}{\parbox[c]{40mm}{\centering V $\rightarrow$ P(r)\\(Unseen Region and Spectral Band)}} &
  65.54 & 74.92 & 50.53 & 26.25 & 74.05 & 56.85 & 45.01 \\
  HRDA\pubyear{2022} (source-only)\cite{HRDA} & MiT-B5 & & 55.81 & 83.77 & 27.57 & 84.42 & 46.57 & 66.80 & 50.91 & & 79.60 & 84.69 & 36.08 & 2.15 & 70.00 & 56.72 & 39.30 \\
  GeoSA-BaSA\pubyear{2025}\cite{GeoSABaSA} & DINOv2-L & & 80.12 & 96.55 & \textbf{68.96} & 77.05 & \underline{88.74} & 82.57 & 71.16 & & 77.94 & 94.06 & 77.52 & 31.61 & 86.10 & 72.11 & 58.43 \\
  CrossEarth\pubyear{2025}\cite{CrossEarth} & DINOv2-L & & \textbf{85.87} & 97.80 & 55.61 & 85.14 & 80.59 & 84.34 & \underline{73.14} & & 76.81 & \textbf{98.67} & 43.88 & 13.56 & 88.45 & 63.41 & 49.04 \\

  Frozen & DINOv3-L & & 72.04 & 98.87 & 5.93 & 81.77 & 71.66 & 71.81 & 53.87 & & 92.05 & 96.31 & 59.26 & 15.43 & 83.75 & 70.81 & 54.21 \\
  Full Fine-Tune & DINOv3-L & & 53.88 & 99.20 & 21.27 & 92.12 & 70.71 & 72.05 & 54.82 & & 89.93 & 94.49 & 73.65 & \textbf{41.43} & 78.70 & 76.48 & 63.15 \\
  Adapter\pubyear{2019}\cite{Adapter} & DINOv3-L & & 73.40 & 98.90 & 24.73 & \textbf{95.32} & 74.61 & 78.67 & 61.71 & & 85.24 & 93.66 & 81.11 & 35.72 & \underline{88.98} & 75.77 & 63.79 \\
  LoRA\pubyear{2022}\cite{LoRA} & DINOv3-L & & 70.64 & 98.95 & 31.61 & \underline{94.35} & 78.87 & 78.89 & 63.36 & & \underline{93.50} & 95.42 & 81.90 & 21.67 & 80.51 & 76.56 & 62.13 \\
  VPT\pubyear{2022}\cite{VPT} & DINOv3-L & & 69.29 & \textbf{99.37} & 26.09 & 87.41 & 76.22 & 76.00 & 60.64 & & 75.14 & 92.19 & 78.81 & 5.53 & 57.69 & 66.71 & 48.09 \\
  Rein\pubyear{2024}\cite{Rein} & DINOv3-L & & 71.64 & \underline{99.32} & 36.82 & 91.84 & 82.20 & 79.49 & 64.83 & & 82.15 & 97.95 & 77.03 & 34.42 & 87.53 & 74.87 & 62.21 \\
  Earth-Adapter\pubyear{2026}\cite{EarthAdapter} & DINOv3-L & & 82.32 & 97.21 & 36.58 & 91.06 & 78.09 & 81.49 & 66.24 & & 76.57 & 96.86 & \underline{87.33} & 38.79 & 82.88 & 75.69 & 63.58 \\
  MM-DINO\pubyear{2026}\cite{MMDINO} & DINOv3-L & & 80.27 & 98.18 & 63.84 & 85.46 & \textbf{92.69} & \underline{84.46} & 73.01 & & 83.07 & \underline{98.28} & \textbf{89.47} & 38.74 & \textbf{95.90} & \underline{78.57} & \underline{65.78} \\
  \rowcolor{gray!25}
  \textbf{GeoDINO (Ours)} & DINOv3-L & & \underline{84.44} & 96.26 & \underline{63.98} & 90.69 & 79.75 & \textbf{86.20} & \textbf{73.67} & & \textbf{93.73} & 95.16 & 82.31 & \underline{41.35} & 79.77 & \textbf{79.63} & \textbf{66.57} \\

  \end{tabular}%
  }%
  \resizebox{0.97\textwidth}{!}{%
  \begin{tabular}{@{}c@{}}
  \usebox{\dgispbox}\\
  \begin{tabularx}{\wd\dgispbox}{l c >{\centering\arraybackslash}X *{8}{r} r >{\centering\arraybackslash}X *{8}{r} r}
  \midrule
  \multirow{2}{*}[-1.1ex]{Method} &
  \multirow{2}{*}[-1.1ex]{Backbone} &
  \multicolumn{1}{c}{Setting} & \multicolumn{8}{c}{Classes --- Acc} & \multirow{2}{*}[-1.1ex]{mIoU} &
  \multicolumn{1}{c}{Setting} & \multicolumn{8}{c}{Classes --- Acc} & \multirow{2}{*}[-1.1ex]{mIoU} \\
  \cmidrule(lr){3-3}\cmidrule(lr){4-11}\cmidrule(lr){13-13}\cmidrule(lr){14-21}
  & & Train $\rightarrow$ Test & Crop & Forest & Grass & Wetl & Water & Unus & Built & Total &  &
  Train $\rightarrow$ Test & Crop & Forest & Grass & Wetl & Water & Unus & Built & Total &  \\
  \midrule
  DAFormer\pubyear{2022} (source-only)\cite{DAFormer} & MiT-B5 &
  \multirow{13}{*}{\parbox[c]{24mm}{\centering H(W) $\rightarrow$ H(E)\\(Unseen Region)}} &
  63.53 & 79.19 & 19.35 & 0.74 & 65.89 & 28.32 & 80.82 & 66.87 & 35.72 &
  \multirow{13}{*}{\parbox[c]{24mm}{\centering H(E) $\rightarrow$ H(W)\\(Unseen Region)}} &
  60.75 & 87.24 & 31.81 & 12.38 & 65.01 & 33.31 & \textbf{78.91} & 72.56 & 41.03 \\
  HRDA\pubyear{2022} (source-only)\cite{HRDA} & MiT-B5 & & 59.22 & \textbf{86.71} & 15.47 & \underline{7.02} & 57.14 & 31.54 & \underline{81.67} & 64.86 & 35.31 & & 60.18 & 87.84 & 28.07 & 15.23 & 65.67 & 34.88 & \underline{77.51} & 72.33 & 40.99 \\
  GeoSA-BaSA\pubyear{2025}\cite{GeoSABaSA} & DINOv2-L & & 65.35 & \underline{86.08} & 17.96 & 0.65 & 67.04 & \textbf{45.73} & 73.57 & 68.32 & 37.07 & & 56.28 & 87.18 & 27.47 & 28.64 & 79.69 & 36.77 & 74.72 & 72.02 & 42.67 \\
  CrossEarth\pubyear{2025}\cite{CrossEarth} & DINOv2-L & & 64.98 & 77.77 & 14.99 & 2.24 & 53.44 & 25.37 & 69.20 & 61.77 & 31.51 & & 56.18 & 81.84 & 38.74 & 16.42 & 64.91 & 30.92 & 73.95 & 69.12 & 38.32 \\

  Frozen & DINOv3-L & & 72.39 & 80.72 & 15.83 & 0.11 & 75.65 & 29.86 & \textbf{83.34} & 72.97 & 39.68 & & \underline{66.84} & 88.65 & 34.68 & 23.49 & 69.39 & 32.05 & 73.80 & 74.74 & 45.13 \\
  Full Fine-Tune & DINOv3-L & & 71.12 & 81.86 & 23.89 & 0.40 & 79.23 & 27.93 & 79.22 & 73.32 & 40.56 & & 61.40 & 87.87 & \textbf{45.77} & 23.98 & 75.00 & 36.79 & 76.17 & 75.02 & 46.01 \\
  Adapter\pubyear{2019}\cite{Adapter} & DINOv3-L & & 67.60 & 81.35 & \underline{24.75} & 0.63 & 77.21 & 35.30 & 80.02 & 71.81 & 40.06 & & 61.26 & 88.73 & 36.71 & 23.47 & 70.17 & 35.60 & 75.96 & 74.15 & 43.94 \\
  LoRA\pubyear{2022}\cite{LoRA} & DINOv3-L & & \textbf{75.37} & 80.19 & 23.04 & 0.43 & 80.94 & 32.67 & 71.67 & \underline{73.74} & 40.62 & & 63.99 & 88.02 & 39.40 & 13.46 & \textbf{83.40} & 33.60 & 76.74 & 75.44 & 45.59 \\
  VPT\pubyear{2022}\cite{VPT} & DINOv3-L & & \underline{74.44} & 79.79 & 9.79 & 0.10 & 74.07 & 31.38 & 81.08 & 72.56 & 39.06 & & 56.25 & 88.46 & \underline{43.80} & 27.19 & 72.75 & \textbf{42.96} & 75.24 & 73.94 & 45.10 \\
  Rein\pubyear{2024}\cite{Rein} & DINOv3-L & & 74.00 & 77.34 & \textbf{28.86} & 0.03 & \underline{82.13} & 32.69 & 73.45 & 73.43 & \underline{41.20} & & 57.94 & \textbf{92.57} & 30.63 & \textbf{32.02} & 81.88 & \underline{38.91} & 73.15 & \underline{75.51} & \underline{47.31} \\
  Earth-Adapter\pubyear{2026}\cite{EarthAdapter} & DINOv3-L & & 70.98 & 84.42 & 14.98 & 1.05 & 78.01 & 32.78 & 73.38 & 72.53 & 39.83 & & 64.07 & 86.45 & 43.58 & 10.81 & 78.77 & 36.69 & 77.20 & 74.80 & 44.86 \\
  MM-DINO\pubyear{2026}\cite{MMDINO} & DINOv3-L & & 69.91 & 74.99 & 11.43 & 1.67 & 76.98 & 38.34 & 80.53 & 71.08 & 39.11 & & 59.12 & \underline{90.10} & 30.29 & 8.52 & 72.76 & 33.31 & 74.66 & 73.62 & 41.65 \\
  \rowcolor{gray!25}
  \textbf{GeoDINO (Ours)} & DINOv3-L & & 66.70 & 84.82 & 12.03 & \textbf{8.22} & \textbf{87.70} & \underline{44.64} & 75.56 & \textbf{74.60} & \textbf{43.06} & & \textbf{68.37} & 88.22 & 39.93 & \underline{30.26} & \underline{82.05} & 37.91 & 69.52 & \textbf{75.96} & \textbf{48.18} \\

  \bottomrule
  \end{tabularx}
  \end{tabular}%
  }
   \end{table*}

\begin{figure*}[t]
  \centering
  \includegraphics[width=0.83\textwidth]{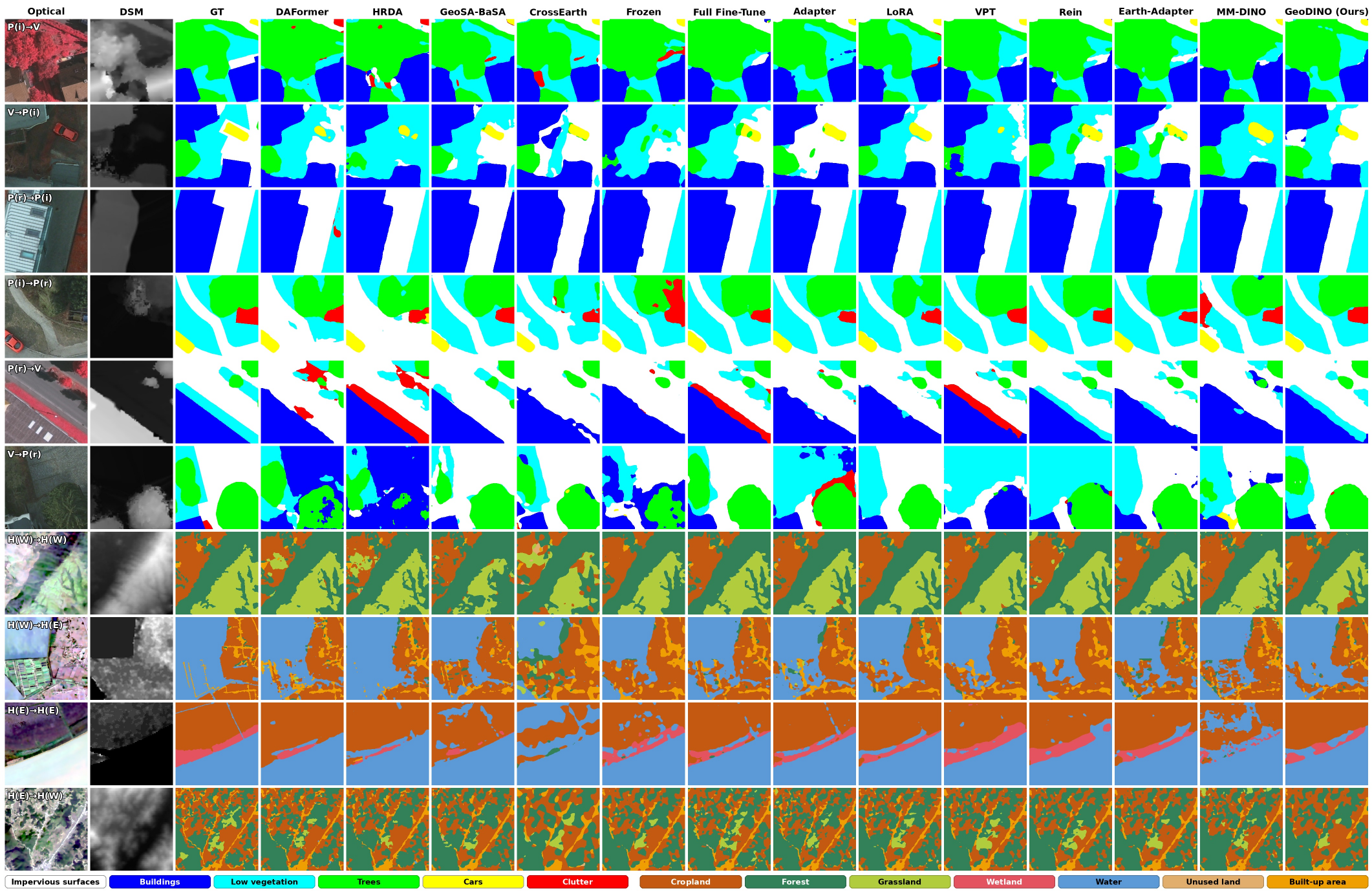}
  \caption{Visualizations of predicted segmentation maps on DG tasks and on the in-domain tasks H(W)$\rightarrow$H(W) and H(E)$\rightarrow$H(E).}
  \label{fig:dg_vis}
\end{figure*}

Table~\ref{tab:dg_6tasks} reports the DG comparison. DAFormer\cite{DAFormer}
and HRDA\cite{HRDA} serve as   source-only DA architecture  baselines,
GeoSA-BaSA\cite{GeoSABaSA}
and CrossEarth\cite{CrossEarth} are included as single-modal RSDG references,
 MM-DINO\mbox{\cite{MMDINO}} is included as a multimodal reference with the same
DINOv3-L backbone,
 and Frozen, Full Fine-Tune, Adapter\cite{Adapter}, LoRA\cite{LoRA},
VPT\cite{VPT}, Rein\cite{Rein}, Earth-Adapter\cite{EarthAdapter}, and
GeoDINO are compared within the same GeoDINO framework to isolate the effect
of the PEFT design. Among all compared methods, GeoDINO achieves the best
mIoU on all eight cross-domain tasks, covering
unseen-region transfer, unseen-spectral transfer, and their combination.

For the unseen-region transfers, GeoDINO delivers the strongest results within
the controlled PEFT comparison, reaching 73.64\% mIoU on P(i)$\rightarrow$V,
outperforming LoRA by 1.69\%, and 71.85\% mIoU on V$\rightarrow$P(i),
surpassing Full Fine-Tune by 1.82\%. These results indicate that
geometry-aware multimodal adaptation is especially effective when the source
and target domains differ in geographic region.

For the unseen-spectral transfers, GeoDINO again ranks first within the same
PEFT comparison. It achieves 85.47\% mIoU on P(r)$\rightarrow$P(i), improving
over Full Fine-Tune by 0.12\%, and 86.96\% mIoU on P(i)$\rightarrow$P(r),
exceeding Earth-Adapter by 0.17\%. Although the margins are smaller than those
in the unseen-region setting, the proposed framework remains stable under
spectral variation.

The advantage becomes more pronounced under the joint region--spectral shift
than under the spectral shift alone.
GeoDINO achieves 73.67\% mIoU on P(r)$\rightarrow$V, surpassing CrossEarth by
0.53\%, and 66.57\% mIoU on V$\rightarrow$P(r), outperforming   MM-DINO  by
  0.79\%. These results indicate that the
proposed geometry-aware adaptation is also beneficial when domain
discrepancy involves substantial geographic mismatch together with spectral
change.

On the two regional transfers of MMHunan in the last block of
Table~\ref{tab:dg_6tasks}, GeoDINO achieves the best OA and the best mIoU on
both, outperforming Rein in mIoU by 1.86\% on H(W)$\rightarrow$H(E) and by
0.87\% on H(E)$\rightarrow$H(W).

Fig.~\ref{fig:dg_vis} further presents visual comparisons on the six
ISPRS cross-domain tasks and on the regional split of MMHunan. Compared with the competing
methods, GeoDINO produces cleaner segmentation maps with
sharper boundaries and more coherent region layouts. Its advantage is
particularly evident in challenging structures such as narrow
clutter regions and boundaries between buildings, trees, and low vegetation,
where other methods often suffer from fragmented predictions or cross-class
confusion. These qualitative results are consistent with the quantitative
gains in Table~\ref{tab:dg_6tasks}, showing that GeoDINO better preserves
structural integrity and suppresses noisy cross-modal correspondences under
domain shift.

Across the six ISPRS cross-domain tasks, GeoDINO yields larger
gains in settings involving region shift than in those with
spectral shift alone, suggesting that DSM-derived geometric
cues remain comparatively stable across domains and are
especially useful when geographic mismatch is present.
\subsection{Ablation Study}
 \label{sec:ablation}

\subsubsection{Effectiveness of Main Components}
\begin{figure}[t]
  \centering
  \includegraphics[width=\columnwidth]{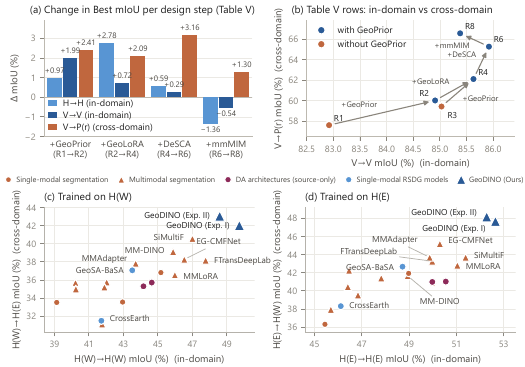}
  \caption{In-domain versus cross-domain effect of the GeoDINO design.
  (a) Change in mIoU for each design step of Table~\ref{tab:ablation_main} on
  H$\rightarrow$H, V$\rightarrow$V, and V$\rightarrow$P(r).
  (b) V$\rightarrow$V versus V$\rightarrow$P(r) mIoU of the configurations in
  Table~\ref{tab:ablation_main}; arrows follow the design steps.
  (c) and (d) In-domain versus cross-domain mIoU on the regional split of MMHunan
  for models trained on H(W) and on H(E), covering the methods of
  Table~\ref{tab:mmhunan} and the source-only DA architectures and
  single-modal RSDG models of Table~\ref{tab:dg_6tasks}; GeoDINO is shown as
  trained in Experiment~I and in Experiment~II, and methods published in
  2025--2026 are named.}
  \label{fig:id_ood}
\end{figure}

We study the effectiveness of the main components in GeoDINO, including
GeoPrior/GeoLoRA, DeSCA, and mmMIM, as summarized in
Table~\ref{tab:ablation_main} and Fig.~\ref{fig:id_ood}. The H$\rightarrow$H and
V$\rightarrow$V columns measure the in-domain effect of each component, and the
V$\rightarrow$P(r) column measures its cross-domain effect.

GeoPrior consistently improves performance under both in-domain and
cross-domain evaluation.
On top of the frozen baseline, adding GeoPrior improves mIoU by 0.97\% on
H$\rightarrow$H, 1.99\% on V$\rightarrow$V, and 2.41\% on
V$\rightarrow$P(r) in R1$\rightarrow$R2; since the encoder is frozen in both
rows, these gains come from the geometry-aware attention bias alone.
A similar trend holds with LoRA-based adaptation, where R4
outperforms R3 by 0.58\% on H$\rightarrow$H, 0.60\% on V$\rightarrow$V, and 2.68\% on
V$\rightarrow$P(r).
Standard LoRA provides a stronger trainable PEFT baseline than the
frozen setting, while GeoLoRA brings additional gains beyond vanilla
PEFT.
Replacing the frozen baseline with standard LoRA in R1$\rightarrow$R3 improves
mIoU by 2.11\% on V$\rightarrow$V and 1.82\% on V$\rightarrow$P(r), whereas
with GeoPrior the same adaptation in R2$\rightarrow$R4 adds 2.78\% on
H$\rightarrow$H, 0.72\% on V$\rightarrow$V, and 2.09\% on V$\rightarrow$P(r).
This controlled comparison between R3 and R4 shows that the improvement
does not merely come from enabling PEFT, but from injecting an explicit
geometry-aware bias, which lets the adaptation gain carry over to the unseen
domain. Along the design steps of Fig.~\ref{fig:id_ood}(a), the
V$\rightarrow$P(r) score rises by 8.96\% from R1 to R8 and the V$\rightarrow$V
score by 3.00\% from R1 to R6. GeoPrior and GeoLoRA contribute 27\% and 23\% of
this cross-domain gain and 66\% and 24\% of this in-domain gain, so the
geometry-guided adaptation has the largest in-domain effect and accounts for
half of the cross-domain gain.

DeSCA improves over the DFM baseline mainly under domain shift. Replacing DFM with
DeSCA under Std-LoRA improves mIoU by 0.25\% on H$\rightarrow$H and 5.06\% on
V$\rightarrow$P(r) from R3 to R5. Under GeoLoRA, the same replacement
yields gains of 0.59\% on H$\rightarrow$H, 0.29\% on V$\rightarrow$V, and 3.16\% on
V$\rightarrow$P(r) from
R4 to R6, suggesting that DeSCA is a stronger fusion module than DFM.
DeSCA contributes 35\% of the cross-domain gain, the largest share of a single
component, but only 10\% of the in-domain gain, which indicates that
its selective sampling matters mainly in an unseen domain, where the relation between the optical
image and the DSM is not always the same as in the source domain.

The mmMIM objective provides an additional gain in the cross-domain
setting: R8 improves over R6 by 1.30\% on
V$\rightarrow$P(r).
The mmMIM objective thus contributes the remaining 15\% of the cross-domain gain, whereas the
in-domain scores decrease by 1.36\% on H$\rightarrow$H and
0.54\% on V$\rightarrow$V in R6$\rightarrow$R8, so mmMIM benefits cross-domain
performance only and acts as a regularizer rather than as a means of fitting
the training distribution.
The full model in R8 achieves the best cross-domain result, showing that
mmMIM offers complementary regularization beyond geometry-aware
adaptation and feature fusion.

\subsubsection{Ablation of Multimodal Feature Fusion}
We further analyze the proposed multimodal fusion design in
Table~\ref{tab:ablation_main}.

Replacing the default DFM baseline with global cross-attention already
improves performance, increasing the score from 54.48\% to 54.56\% on
H$\rightarrow$H and from 62.11\% to 64.38\% on V$\rightarrow$P(r) in
R4$\rightarrow$R9, which confirms the benefit of explicit cross-modal
interaction over simple channel reweighting.
However, the full DeSCA in R6 still performs better than global
cross-attention, by 0.51\% on H$\rightarrow$H and 0.89\% on V$\rightarrow$P(r), suggesting that
dense all-to-all matching is less suitable for optical--DSM fusion under
domain shift.

\setcounter{topnumber}{1}
\begin{table}[t]
\centering
\scriptsize
\caption{Ablation study of GeoDINO on H$\rightarrow$H, V$\rightarrow$V, and V$\rightarrow$P(r).  All scores are mIoU (\%). Best denotes the highest mIoU obtained for each setting.  R1--R8 evaluate the main components, including GeoLoRA, DeSCA, and mmMIM; R9--R11 analyze the multimodal fusion design; R12--R14 compare different mmMIM masking strategies; and R15--R17 are same-backbone references}

\label{tab:ablation_main}

  \setlength{\tabcolsep}{2.4pt}
 \renewcommand{\arraystretch}{1.0}

\begin{tabular*}{\columnwidth}{@{\extracolsep{\fill}}l c c c c}
\toprule
Setting & PEFT & H$\rightarrow$H & V$\rightarrow$V & V$\rightarrow$P(r) \\
\midrule

\rowcolor{gray!25}
\multicolumn{5}{l}{$\blacktriangledown$ Effectiveness of each main component (GeoLoRA, DeSCA, mmMIM)}\\

R1~~Frozen + DFM\cite{MFNet}
& Frozen & 50.73 & 82.92 & 57.61 \\

R2~~R1 + GeoPrior
& Frozen & 51.70 & 84.91 & 60.02 \\

R3~~R1 + LoRA
& Std-LoRA & 53.90 & 85.03 & 59.43 \\

R4~~R3 + GeoPrior
& GeoLoRA & 54.48 & 85.63 & 62.11 \\

R5~~R3 + DeSCA
& Std-LoRA & 54.15 & 84.87 & 64.49 \\

R6~~R4 + DeSCA
& GeoLoRA & 55.07 & 85.92 & 65.27 \\

R7~~R5 + mmMIM
& Std-LoRA & 52.86 & 83.43 & 62.13 \\

R8~~R6 + mmMIM
& GeoLoRA & 53.71 & 85.38 & 66.57 \\

\rowcolor{gray!15}
\multicolumn{5}{l}{$\blacktriangledown$ Ablation of multimodal feature fusion}\\

R9~~R4 + Global CrossAttention
& GeoLoRA & 54.56 & -- & 64.38 \\

R10~~R6 w/o deformable sampling
& GeoLoRA & 54.39 & -- & 63.94 \\

R11~~R6 w/o modality dominance
& GeoLoRA & 54.33 & -- & 64.21 \\

\rowcolor{gray!15}
\multicolumn{5}{l}{$\blacktriangledown$ Ablation of mmMIM}\\

R12~~R6 + mmMIM (both-modal mask)
& GeoLoRA & -- & -- & 65.18 \\

R13~~R6 + mmMIM (optical-only mask)
& GeoLoRA & -- & -- & 66.21 \\

R14~~R6 + mmMIM (DSM-only mask)
& GeoLoRA & -- & -- & 65.86 \\

\rowcolor{gray!15}
\multicolumn{5}{l}{$\blacktriangledown$ Same-backbone references (DINOv3-L, UNetFormer decoder)}\\

R15~~Optical only ($v/o$/FFN LoRA)
& Std-LoRA & 48.56 & 84.57 & 56.43 \\

R16~~R3 + concat.\ fusion
& Std-LoRA & 50.12 & 85.10 & 62.75 \\

R17~~R3 + Global CrossAttention
& Std-LoRA & 50.29 & 84.77 & 63.25 \\

\bottomrule
\end{tabular*}
\end{table}

Removing deformable sampling from DeSCA reduces the score from 55.07\%
to 54.39\% on H$\rightarrow$H and from 65.27\% to 63.94\% on V$\rightarrow$P(r)
in R6$\rightarrow$R10.
This degradation indicates that adaptive sparse sampling helps establish
more reliable cross-modal correspondences than fixed dense interactions.
Removing modality dominance weighting also degrades performance to
54.33\% on H$\rightarrow$H and 64.21\% on V$\rightarrow$P(r) in R11, showing that
dynamic point-wise modality weighting is also important for suppressing
noisy cross-modal interactions.
Taken together, these results suggest that the advantage of DeSCA comes
from the joint effect of deformable sampling and modality dominance
weighting, rather than from sparse cross-attention alone.

\subsubsection{Ablation of Multimodal MIM}
We further analyze the masking strategy in the proposed mmMIM,
as summarized in Table~\ref{tab:ablation_main}.

Compared with the model without mmMIM in R6,   the single-modality masking
 variants improve cross-domain performance,   whereas masking both modalities
simultaneously in R12 does not. Random  single-modality masking achieves the
best result,   supporting the design choice of keeping one reliable stream for
the segmentation loss in each sample.
Among the two single-modality variants, masking the optical
modality performs better than masking DSM, likely because
masking DSM also removes the geometric signal used to
construct the geometry prior for attention guidance.
Overall, masking a randomly chosen modality in each iteration yields the
strongest cross-domain performance among the masking strategies.

\begin{figure*}[t]
  \noindent%
  \begin{minipage}[t]{0.273\textwidth}
  \centering
  \makebox[\linewidth][l]{\hspace*{26.7pt}\includegraphics[height=1.8in]{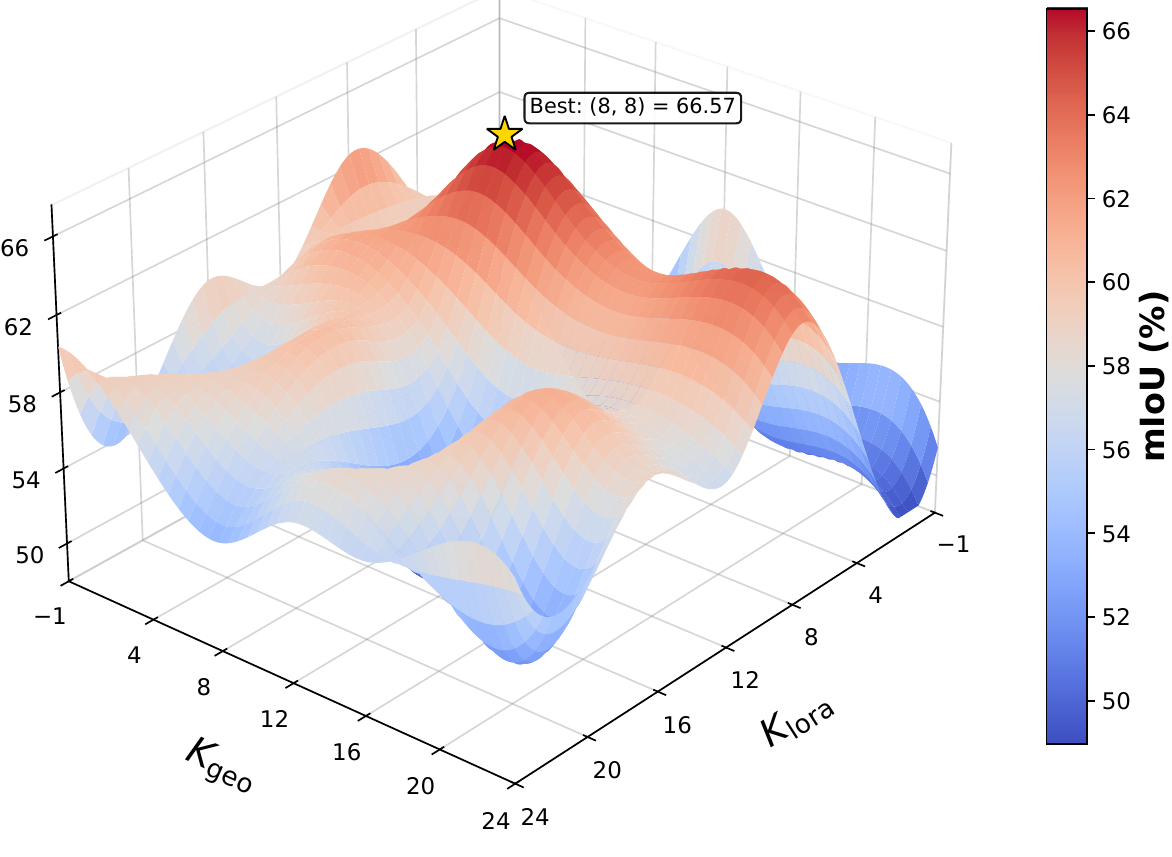}}
  \caption{mIoU (\%) on V$\rightarrow$P(r) as a function of the
  number of final layers with GeoPrior injection ($K_{\mathrm{geo}}$)
  and LoRA adaptation ($K_{\mathrm{lora}}$). The star marks the best
  configuration.
  A value of $-1$ denotes that the corresponding module is disabled.}
  \label{fig:hyperparameter}
  \end{minipage}\hfill
  \begin{minipage}[t]{0.700\textwidth}
  \centering
  \makebox[\linewidth][l]{\hspace*{64.3pt}\includegraphics[height=1.8in]{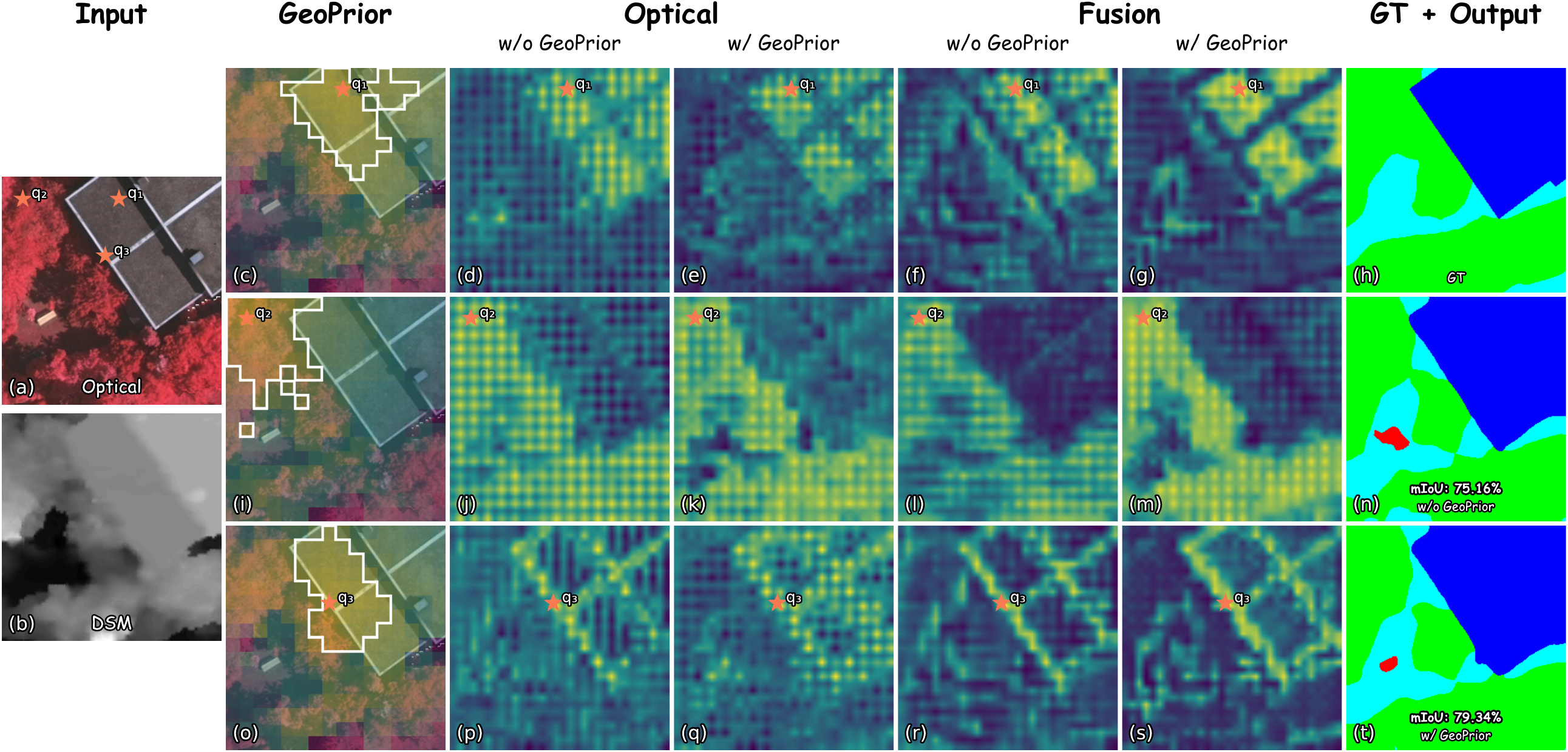}}
  \caption{
Mechanism analysis of GeoPrior on P(i) $\rightarrow$ V.
(a) Optical image with query points. (b) DSM.
(c), (i), and (o) GeoPrior responses for $q_1$, $q_2$, and $q_3$.
(d), (j), and (p) Optical features without GeoPrior.
(e), (k), and (q) Optical features with GeoPrior.
(f), (l), and (r) Fused features without GeoPrior.
(g), (m), and (s) Fused features with GeoPrior.
(h) Ground truth. (n) Prediction without GeoPrior.
(t) Prediction with GeoPrior.
$q_1$, $q_2$, and $q_3$ denote a building-centered query,
a tree-centered query, and a building--tree boundary query, respectively.
White contours denote the top-response support regions in the
GeoPrior maps.
}
  \label{fig:geoprior_vis}
  \end{minipage}
  \par\vspace{-2.5pt}%
\end{figure*}

\subsubsection{Sensitivity to Injection Depth}
A key design choice in GeoDINO is the number of transformer layers
at the end of the backbone that receive geometry-aware modulation
($K_{\mathrm{geo}}$) and LoRA adaptation ($K_{\mathrm{lora}}$).
Fig.~\ref{fig:hyperparameter} reports mIoU on V$\rightarrow$P(r)
across a $7\times7$ grid.

The best mIoU of 66.57\% occurs at $K_{\mathrm{geo}}=8$ and
$K_{\mathrm{lora}}=8$, meaning that GeoPrior injection and LoRA
adaptation are both applied to the last eight layers of the 24-layer
DINOv3-L encoder. The performance surface is relatively flat within a
moderate band ($K\in[8,12]$) but drops at both extremes: using too
few final layers limits adaptation capacity, whereas extending the
modules to nearly all layers disturbs the general pretrained
representations preserved in the shallow stages. Between the two axes,
$K_{\mathrm{lora}}$ has a stronger influence. Most row-wise optima
align at $K_{\mathrm{lora}}=8$, whereas the surface varies more
gently along $K_{\mathrm{geo}}$. This indicates that the number of
final LoRA-adapted layers is the more sensitive hyperparameter.

\subsection{Mechanism Analysis of GeoPrior}
\label{sec:mechanism_analysis}

To better understand how GeoPrior benefits multimodal segmentation across
different evaluation settings, we visualize query-conditioned GeoPrior
responses together with the corresponding optical and fused feature maps, as
shown in Fig.~\ref{fig:geoprior_vis}. We further compare the predictions
without and with GeoPrior to link the intermediate representation changes with
the final segmentation behavior in a typical case.

The visualizations show that GeoPrior acts as a coarse geometry-aware support
cue rather than a boundary-precise mask. For the building-centered query
$q_1$, the GeoPrior response in Fig.~\ref{fig:geoprior_vis}(c) mainly covers
spatially contiguous roof patches that are geometrically consistent with the
query. For the tree-centered query $q_2$, Fig.~\ref{fig:geoprior_vis}(i) shows
that GeoPrior concentrates on the vegetation-side support region while avoiding
most roof areas. For the boundary query $q_3$, the response in
Fig.~\ref{fig:geoprior_vis}(o) is biased toward the building-side support
region rather than expanding across the adjacent tree area, which suppresses
part of the cross-boundary ambiguity in building--tree transition
regions.
Beyond this example, we train two gradient-boosted decision-tree classifiers to predict
the class of $16\times16$-px patches of the Vaihingen training tiles, one
from the DSM height and roughness of each patch and one from its optical
channel means. The DSM classifier is less accurate in-domain but retains 78\%
of its in-domain mF1 on Potsdam, against 26\% for the optical classifier, or
48\% with matched spectral bands. The class evidence carried by DSM geometry
thus transfers better than that of optical appearance.

Consistent with the design in Section~\ref{sec:geo_lora}, GeoPrior is injected
through GeoLoRA into the optical encoder, where it biases the attention toward
geometry-consistent regions. Compared with the variant without GeoPrior, the
optical features with GeoPrior become more compact and less prone to
cross-region leakage, as shown in
Fig.~\ref{fig:geoprior_vis}(e), (k), and (q) versus (d), (j), and (p). This
effect is further reflected in the fused features, where the responses in
Fig.~\ref{fig:geoprior_vis}(g), (m), and (s) exhibit clearer structural
organization and sharper geometry-aligned transitions than those in
(f), (l), and (r). In particular, the improvement is most evident for the
boundary query $q_3$, where GeoPrior helps preserve the building-side
structure while reducing undesired spreading toward the neighboring vegetation
region.

Finally, the prediction comparison in Fig.~\ref{fig:geoprior_vis}(n) and (t),
together with the ground truth in (h), shows that these representation changes
lead to better segmentation behavior. With GeoPrior, the predicted regions are
better aligned with the underlying building--vegetation layout and contain
fewer local errors than the variant without GeoPrior, which supports our
design intuition that GeoPrior first regularizes the optical branch and then
indirectly improves multimodal fusion.
In Table~\ref{tab:ablation_main}, adding GeoPrior to R1, R3, R5, and R7,
which gives R2, R4, R6, and R8, raises the mIoU by 0.60\%--1.99\% on
V$\rightarrow$V and by 0.78\%--4.44\% on V$\rightarrow$P(r): the accuracy
improves on the source domain and on the unseen domain in all four cases, and more on the unseen
domain in three of them.

\begin{figure}[t]
  \centering
  \includegraphics[width=\columnwidth]{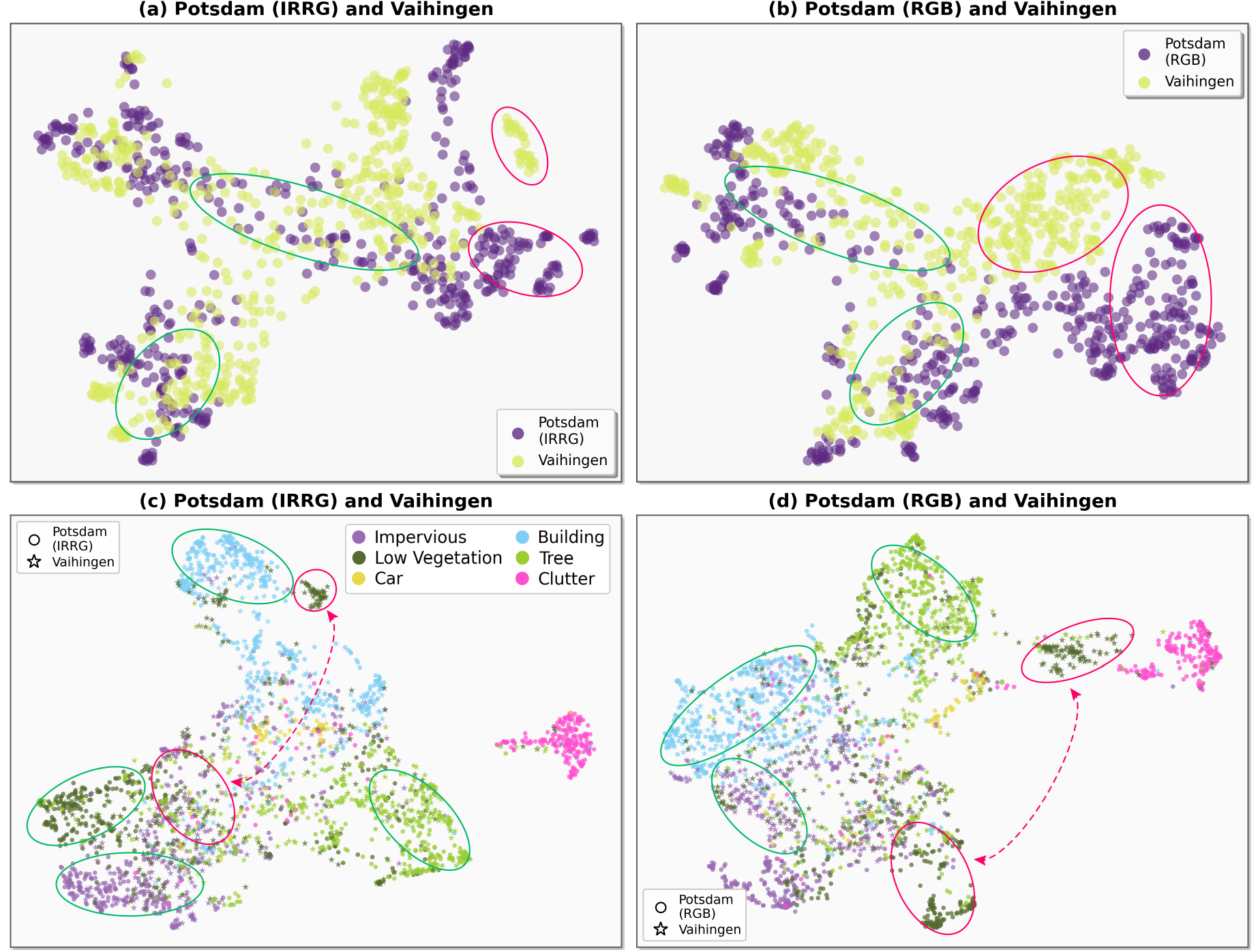}
  \caption{
    (a) and (b) UMAP visualizations of cross-domain feature distributions. Green
    circles mark regions where features from the two domains are closely
    aligned, indicating a smaller domain gap, whereas red circles denote more
    separated distributions and thus a larger gap.
    (c) and (d) UMAP visualizations of interdomain feature clustering. Dots and
    stars denote samples from the two domains, respectively. Green circles
    highlight regions where cross-domain features form compact class-wise
    clusters, suggesting better category alignment. Red circles mark regions
    with substantial overlap or inconsistent separation, indicating categories
    that remain difficult to distinguish across domains.
    }
  \label{fig:DomainGap}
\end{figure}

\begin{figure}[t]
  \centering
  \includegraphics[width=\columnwidth]{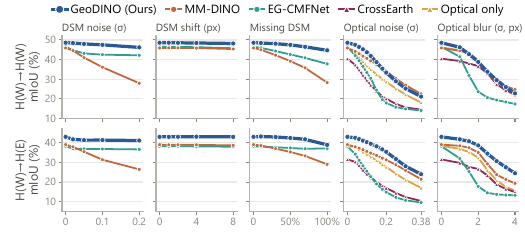}
  \caption{Stress tests on the regional split of MMHunan with H(W) as the
  training part: mIoU against the level of each perturbation on
  H(W)$\rightarrow$H(W) (top) and H(W)$\rightarrow$H(E) (bottom). Level 0 is
  the clean input. CrossEarth and the optical-only DINOv3 with LoRA do not use
  the DSM.}
  \label{fig:stress}
\end{figure}

\subsection{Visualization Analysis}
\subsubsection{Domain Gaps}
The UMAP projections in Fig.~\ref{fig:DomainGap}(a) and~(b),
computed from the last backbone layer using the UMAP
technique~\cite{UMAP}, show that Potsdam~(IRRG) is better aligned with
Vaihingen than Potsdam~(RGB). In Fig.~\ref{fig:DomainGap}(a), samples
from the two domains are more strongly intermixed, indicating a
relatively small domain gap. In Fig.~\ref{fig:DomainGap}(b), by
contrast, more domain-specific subclusters remain, revealing a larger
feature-level gap. Since Vaihingen itself follows an IRRG spectral
configuration, this pattern   illustrates the feature-level effect of spectral
mismatch. The quantitative comparison of the transfer settings is reported in
 Table~\ref{tab:dg_6tasks}.

\subsubsection{Cross-Domain Representation Ability}
Fig.~\ref{fig:DomainGap}(c) and~(d), derived from the fused
features $f_{\text{fuse}}$, illustrate the quality of the
cross-domain representations learned by GeoDINO after
multimodal interaction. Buildings and clutter form compact,
relatively well-separated clusters in both settings.
Same-class samples from the two domains align more tightly
under P(i)$\rightarrow$V in Fig.~\ref{fig:DomainGap}(c) than
under P(r)$\rightarrow$V in Fig.~\ref{fig:DomainGap}(d),
consistent with the smaller domain gap observed in the top
row. Noticeable overlaps remain among impervious surfaces,
low vegetation, and trees, while car features are sparse due
to the limited sample count. These residual overlaps reflect
the intrinsic visual similarity of these categories, which
remains challenging even after fusion.

\begin{table}[t]
\centering
\scriptsize
\caption{Model scale analysis of selected methods on the V$\rightarrow$P(r) task with an input size of $256\times256$, measured on a single NVIDIA GeForce RTX 4090 GPU. The compared methods include single-modal in-domain baselines, multimodal in-domain baselines, DA models, single-modal RSDG references, and same-framework PEFT variants. FLOPs, trainable parameters, memory, inference speed, and mIoU are reported}
\label{tab:model_scale}

\setlength{\tabcolsep}{3.5pt}
\renewcommand{\arraystretch}{1.05}
  \resizebox{\columnwidth}{!}{
\begin{tabular}{l l c c c c c}
\toprule
\textbf{Method} &
\textbf{Backbone} &
\makecell{\textbf{FLOPs(G)}} &
\makecell{\textbf{Param.(M)}} &
\makecell{\textbf{Memory(MB)}} &
\makecell{\textbf{Speed(FPS)}} &
\makecell{\textbf{mIoU(\%)}} \\
\midrule
FuseNet~\cite{FuseNet}              & VGG16         & 120.35 & 44.17  & 297.73 & \textbf{164.76} & 31.70 \\
PSPNet~\cite{PSPNet}                     & ResNet101     & 139.56 & 72.57  & 384.44 & \underline{70.44} & 31.00 \\
TransUNet~\cite{TransUNet}               & R50-ViT-B     & 76.88  & 105.32 & 454.60  & 47.10 & 42.96 \\
FTUNetFormer~\cite{UNetFormer}           & Swin-B        & 70.98  & 95.98  & 414.50  & 35.07 & 43.00 \\
CMGFNet~\cite{CMGFNet}              & ResNet34      & \underline{20.45} & 43.68  & 238.19 & 57.16 & 38.57 \\
CMFNet~\cite{CMFNet}                & VGG16         & 158.72 & 104.07 & 611.94 & 20.78 & 20.93 \\
ESANet~\cite{ESANet}                & ResNet34      & \textbf{19.13} & 46.10  & \textbf{207.98} & 43.86 & 28.54 \\
FTransUNet~\cite{FTransUNet}        & R50-ViT-B     & 139.25 & 203.40 & 829.76 & 22.76 & 48.26 \\
FTransDeepLab~\cite{FTransDeepLab}  & SegFormer-B2  & 37.10  & 60.95  & 456.55 & 25.94 & 51.54 \\
SiMultiF~\cite{SiMultiF}            & Swin-T        & 38.41  & 42.90  & \underline{225.15} & 45.62 & 45.71 \\
EG-CMFNet~\cite{EGCMFNet}           & BiFormer-T    & 38.24  & 53.61  & 302.53 & 20.62 & 42.79 \\
MMLoRA~\cite{MFNet}                 & SAM-L         & 326.76 & \underline{8.98} & 1269.42 & 17.68 & 56.69 \\
MMAdapter~\cite{MFNet}              & SAM-L         & 353.10 & 56.68  & 1454.45 & 9.91 & 57.66 \\
MM-DINO~\cite{MMDINO}               & DINOv3-L      & 388.72 & 31.20  & 1394.01 & 15.07 & \underline{60.32} \\
GeoDINO (Ours)                        & DINOv3-L         & 325.22 & \textbf{6.82} & 1417.52 & 20.12 & \textbf{65.27} \\
\midrule
DAFormer (source-only)~\cite{DAFormer} & MiT-B5       & \textbf{59.34} & 85.15  & \textbf{436.85} & 20.56 & 45.01 \\
HRDA (source-only)~\cite{HRDA}         & MiT-B5       & \underline{75.74} & 89.86  & \underline{457.61} & 10.19 & 39.30 \\
GeoSA-BaSA~\cite{GeoSABaSA}                     & DINOv2-L         & 421.28 & 36.74 & 1392.73 & 5.42 & 58.43 \\
CrossEarth~\cite{CrossEarth}                    & DINOv2-L         & 204.25 & 23.57 & 1325.78 & 10.63 & 49.04 \\
Frozen                                          & DINOv3-L         & 324.40 & \textbf{10.69} & 1433.11 & \underline{23.95} & 54.21 \\
Full Fine-Tune                         & DINOv3-L         & 324.40 & 313.82 & 1433.11 & \textbf{24.76} & 63.15 \\
Adapter~\cite{Adapter}                 & DINOv3-L         & 327.66 & 17.03 & 1511.58 & 17.92 & 63.79 \\
LoRA~\cite{LoRA}                       & DINOv3-L         & 326.15 & 13.84 & 1445.11 & 16.63 & 62.13 \\
VPT~\cite{VPT}                         & DINOv3-L         & 334.07 & \underline{11.08} & 1438.11 & 21.95 & 48.09 \\
Rein~\cite{Rein}                       & DINOv3-L         & 365.34 & 13.65 & 1446.16 & 15.93 & 62.21 \\
Earth-Adapter~\cite{EarthAdapter}     & DINOv3-L         & 329.02 & 20.28 & 1469.64 & 18.62 & 63.58 \\
MM-DINO~\cite{MMDINO}                  & DINOv3-L         & 388.72 & 31.20 & 1394.01 & 15.17 & \underline{65.78} \\
GeoDINO (Ours)                        & DINOv3-L         & 324.95 & 11.61 & 1436.56 & 20.53 & \textbf{66.57} \\
\bottomrule
\end{tabular}
}
 \end{table}

\subsection{Stress Tests Under Input Perturbations}

The models trained on H(W) are further evaluated with perturbed test inputs.
Following \mbox{HDSMNet}\mbox{\cite{HDSMNet}}, all perturbations
are applied at inference without retraining, and results with random
perturbations are averaged over three seeds. For the DSM, Gaussian noise with
a standard deviation $\sigma$ of 0.02--0.20 of the normalized height range is added; the DSM is shifted by 1--8 pixels, or blocks that cover 10\%--70\% of the DSM, up to the entire DSM, are removed. For the optical image,
Gaussian noise with $\sigma$ from 0.04 to 0.38 on the $[0,1]$ image scale is
added, or a Gaussian blur with $\sigma$ from 1 to 4 pixels is applied.

These stress tests demonstrate that GeoDINO is robust to a noisy, shifted, or
missing DSM: its mIoU decreases by at most 4.09\% on both
H(W)$\rightarrow$H(W) and H(W)$\rightarrow$H(E), whereas that of MM-DINO
decreases by up to 17.95\%. As shown in Fig.~\ref{fig:stress}, GeoDINO retains
the highest mIoU on H(W)$\rightarrow$H(W) at every level of DSM noise and missing DSM, and on H(W)$\rightarrow$H(E) at every level of DSM noise, missing
DSM, optical noise, and optical blur. Its advantage on the unseen region
therefore holds under degraded test inputs.

\setcounter{topnumber}{2}%
\subsection{Parameter Efficiency Analysis}

We further compare model scale and efficiency in Table~\ref{tab:model_scale},
distinguishing parameter efficiency, measured by the number of updated
parameters, from computational cost, measured by FLOPs, memory, and throughput.
In the first group, GeoDINO achieves the best mIoU of 65.27\% with only 6.82M trainable parameters, the smallest count in the table, whereas MM-DINO,
MMAdapter, and FTransDeepLab require 31.20M, 56.68M, and 60.95M parameters.
GeoDINO still surpasses MM-DINO and MMAdapter by 4.95\% and 7.61\% mIoU.

In the second group, GeoDINO is compared with MiT-B5 DA models,
DINOv2-L single-modal RSDG models, DINOv3-L PEFT baselines, and MM-DINO.
GeoDINO again attains the best mIoU, reaching 66.57\% with 11.61M
trainable parameters, whereas GeoSA-BaSA and CrossEarth require
36.74M and 23.57M parameters, MM-DINO 31.20M, Adapter 17.03M, and Full
Fine-Tune 313.82M. Its memory footprint remains close to other DINOv3-L PEFT variants;
like these variants, however, its FLOPs and memory reflect a ViT-L encoder
applied to two modalities and exceed those of the MiT-B5 DA models.
Although GeoDINO does not achieve the highest FPS, its inference speed remains
practical. Overall, the results support parameter-efficient adaptation of a
large frozen encoder rather than overall computational efficiency.

\section{Conclusion}
In this study, we present GeoDINO, a  geometry-aware and parameter-efficient
framework for   optical--DSM semantic segmentation that learns from a single
labeled source domain and remains accurate  both in-domain and   on unseen
target domains.
Our results indicate that DSM geometry, used as an explicit prior for adapting a
frozen VFM, raises the accuracy on unseen domains together with the in-domain
accuracy, and that its class evidence transfers better than optical appearance.
GeoDINO matches or exceeds the compared in-domain methods on Vaihingen, Potsdam,
and MMHunan, achieves the best mIoU among the source-only DA architectures,
single-modal RSDG models, and same-framework PEFT variants on eight transfer
settings covering region shift, spectral shift, and their combination, and is
robust to a noisy, shifted, or missing DSM. Its benefit depends on the terrain
and on DSM reconstruction quality, and transfer between different
satellite-scale datasets and sensors remains to be evaluated. We hope that this
work draws attention to auxiliary geometry as a source of generalization for
remote sensing segmentation.

\end{sloppypar}
\end{document}